%% file: 0-main.tex
\documentclass{article}

\usepackage[final]{corl_2021} 

\usepackage{wrapfig}

\usepackage[utf8]{inputenc} 
\usepackage[T1]{fontenc}    
\usepackage{hyperref}       
\usepackage{enumitem}
\usepackage{url}            
\usepackage{booktabs}       
\usepackage[table]{xcolor}    
\hypersetup{
    colorlinks=true,
    linkcolor=blue,
    filecolor=magenta,      
    urlcolor=cyan,
}
\usepackage{amsfonts}       
\usepackage{nicefrac}       
\usepackage{microtype}      
\usepackage{algorithm} 
\usepackage{algorithmicx, algpseudocode} 
\usepackage{mathtools}
\usepackage{amsmath, amssymb, amscd}
\usepackage{amsthm}
\usepackage{wasysym} 
\usepackage{amsfonts}
\usepackage{gensymb} 
\usepackage{nicefrac} 
\usepackage{verbatim}
\usepackage[skip=2pt,font=small]{caption}
\usepackage{subcaption}
\usepackage{bbold}
\usepackage{xspace}
\usepackage{color}
\usepackage{soul} 
\usepackage{xr}
\usepackage{chngcntr} 
\usepackage{siunitx}
\usepackage{multirow}
\usepackage{fontawesome}
\usepackage{placeins}

\definecolor{indiagreen}{rgb}{0.07, 0.53, 0.03}

\newcommand{\States}{\mathcal{S}}

\newcommand{\ig}{iGibson 2.0\xspace}

\newcommand{\semicheckmark}{\checkmark\kern-1.1ex\raisebox{.7ex}{\rotatebox[origin=c]{125}{--}}
}

\usepackage{xr}
\makeatletter
\newcommand*{\addFileDependency}[1]{
  \typeout{(#1)}
  \@addtofilelist{#1}
  \IfFileExists{#1}{}{\typeout{No file #1.}}
}
\makeatother
 
\newcommand*{\inputexternaldocument}[1]{%
    \externaldocument{#1}%
    \addFileDependency{#1.tex}%
    \addFileDependency{#1.aux}%
}

\usepackage{titlesec}
\titlespacing*{\section}{8pt}{8pt plus 4pt minus 4pt}{4pt plus 2pt minus 2pt}
\titlespacing*{\subsection}{4pt}{4pt plus 1pt minus 1pt}{2pt plus 1pt minus 1pt}

\inputexternaldocument{5-appendix}

\title{\LARGE{\ig: Object-Centric Simulation for\\Robot Learning of Everyday Household Tasks}}

\author{%
Chengshu Li*\thanks{indicates equal contribution\newline correspondence to \href{mailto:chengshu@stanford.edu,feixia@stanford.edu,robertom@stanford.edu}{\{chengshu,feixia,robertom\}@stanford.edu}}$^{\scriptscriptstyle \spadesuit}$, Fei Xia*$^{\scriptscriptstyle \heartsuit}$, Roberto Mart\'in-Mart\'in*$^{\scriptscriptstyle \spadesuit}$
\\\textbf{Michael Lingelbach$^{\scriptscriptstyle \clubsuit}$, Sanjana Srivastava$^{\scriptscriptstyle \spadesuit}$, Bokui Shen$^{\scriptscriptstyle \spadesuit}$, Kent Vainio$^{\scriptscriptstyle \spadesuit}$}
\\\textbf{Cem Gokmen$^{\scriptscriptstyle \spadesuit}$, Gokul Dharan$^{\scriptscriptstyle \spadesuit}$, Tanish Jain$^{\scriptscriptstyle \spadesuit}$, Andrey Kurenkov$^{\scriptscriptstyle \spadesuit}$}
\\\textbf{Karen Liu$^{\scriptscriptstyle \spadesuit\bigstar}$, Hyowon Gweon$^{\scriptscriptstyle \diamondsuit\bigstar}$, Jiajun Wu$^{\scriptscriptstyle \spadesuit\bigstar}$, Li Fei-Fei$^{\scriptscriptstyle \spadesuit\bigstar}$, Silvio Savarese$^{\scriptscriptstyle \spadesuit\bigstar}$}
\\\\Department of Computer Science$^{\scriptscriptstyle \spadesuit}$, Electrical Engineering$^{\scriptscriptstyle \heartsuit}$, Neurosciences IDP$^{\scriptscriptstyle \clubsuit}$, Psychology$^{\scriptscriptstyle \diamondsuit}$
\\Institute for Human-Centered AI (HAI)$^{\scriptscriptstyle \bigstar}$
\\Stanford University
}
\begin{document}

\maketitle
\vspace{-25pt}
\begin{abstract}
Recent research in embodied AI has been boosted by the use of simulation environments to develop and train robot learning approaches. However, the use of simulation has skewed the attention to tasks that only require what robotics simulators can simulate: motion and physical contact. We present \ig, an open-source simulation environment that supports the simulation of a more diverse set of household tasks through three key innovations. First, \ig supports object states, including temperature, wetness level, cleanliness level, and toggled and sliced states, necessary to cover a wider range of tasks. Second, \ig implements a set of predicate logic functions that map the simulator states to logic states like \texttt{Cooked} or \texttt{Soaked}. Additionally, given a logic state, \ig can sample valid physical states that satisfy it. This functionality can generate potentially infinite instances of tasks with minimal effort from the users. The sampling mechanism allows our scenes to be more densely populated with small objects in semantically meaningful locations. Third, \ig includes a virtual reality (VR) interface to immerse humans in its scenes to collect demonstrations. As a result, we can collect demonstrations from humans on these new types of tasks, and use them for imitation learning. We evaluate the new capabilities of \ig to enable robot learning of novel tasks, in the hope of demonstrating the potential of this new simulator to support new research in embodied AI. \ig and its new dataset are publicly available at \href{http://svl.stanford.edu/igibson/}{http://svl.stanford.edu/igibson/}.
\end{abstract}

\vspace{-5pt}
\input{1-intro-v2}

\vspace{-5pt}
\input{2-rw}
\vspace{-5pt}
\input{3a-extendedstates}
\vspace{-5pt}
\input{3b-predicates}
\vspace{-5pt}
\input{3c-VR}

\vspace{-5pt}
\input{4-eval}
\vspace{-5pt}
\section{Conclusion}

We presented \ig, an open-source simulation environment for household tasks with several key novel features: 1) an object-centric representation and extended object states (e.g. temperature, wetness and cleanliness level), 2) logical predicates mapping simulation states to logical states, and generative mechanism to create simulated worlds based on a given logical description, and 3) a virtual reality interface to easily collect human demonstrations for imitation. We demonstrate in multiple experiments the new avenues for research enabled by \ig. We hope \ig becomes a useful tool for the community, and facilitates the development of novel embodied AI solutions.

\acknowledgments{
This work is in part supported by ARMY MURI grant W911NF-15-1-0479 and Stanford Institute for Human-Centered AI (SUHAI). 
S. S. is supported by the National Science Foundation Graduate Research Fellowship Program (NSF GRFP) and Department of Navy award (N00014-16-1-2127) issued by the Office of Naval Research. S. S. and C. L. are supported by SUHAI Award \# 202521. R. M-M is supported by SAIL TRI Center – Award \# S-2018-28-Savarese-Robot-Learn. F. X. and B. S. are supported by Qualcomm Innovation Fellowship. M. L. is supported by Regina Casper Stanford Graduate Fellowship.
}


\renewcommand*{\bibfont}{\footnotesize}

\renewcommand{\baselinestretch}{.95}
\bibliography{refs}


\input{5-appendix}

\end{document}

%% file: 1-intro-v2.tex
\section{Introduction}
\label{s:intro}


Recent years, we have seen the emergence of many simulation environments for robotics and embodied AI research~\cite{shen2020igibson,xia2018gibson,xia2020interactive,xiang2020sapien,habitat19arxiv,ehsani2021manipthor,gan2020threedworld,james2019rlbench,zhu2020robosuite, yu2020meta}. 
The main function of these simulators is to compute the motion resulting from the physical contact-interaction between (rigid) bodies, as this is the main process that allows robots to navigate and manipulate the environment.
This kinodynamic simulation is sufficient for pick-and-place and rearrangement tasks~\cite{batra2020rearrangement,weihs2021visual,gan2021threedworld,liu2021ocrtoc}; however, as the field advances, researchers are taking on more diverse and complex tasks that cannot be performed in these simulators, e.g., household activities that involve changing the temperature of objects, their dirtiness and wetness levels.
There is a need for new simulation environments that can maintain and update new types of object states to broaden the diversity of activities that can be studied.

We present \ig, an open-source extension of the kinodynamic simulator iGibson with several novel functionalities. First and foremost, \ig maintains and updates new \textbf{extended physical states} resulting from the approximation of additional physical processes. These states include not only kinodynamics (pose, motion, forces), but also object's temperature, wetness level, cleanliness level, toggled and sliced state (functional states).
These states have a direct effect on the appearance of the objects, captured by the high-quality virtual sensor signals rendered by the simulator.


Second, \ig provides a set of logical predicates that can be evaluated with a single object (e.g. \texttt{Cooked}) or a pair of objects (e.g. \texttt{InsideOf}). These logical predicates discriminate the continuous physical state maintained by the simulator into semantically meaningful logical states (e.g. \texttt{Cooked} is \texttt{True} if the temperature is above a certain threshold).
Complementary to the discriminative functions, \ig implements \textbf{generative functions} that sample valid simulated physical states based on logical states. Scene initialization can then be described as a set of logical states that the simulator can translate into valid physical instances. This enables faster prototyping and specification of scenes in \ig, facilitating the training of embodied AI agents in diverse instances of the same tasks. We demonstrate the potential of this generative functionality with a new dataset of home scenes densely populated with small objects. We generate this new dataset by applying a set of hand-designed semantic-logic rules to the original scenes of iGibson 1.0.

\begin{wrapfigure}{r}{0.6\textwidth}
\vspace{-1em}
    \centering
    \includegraphics[width=0.99\linewidth]{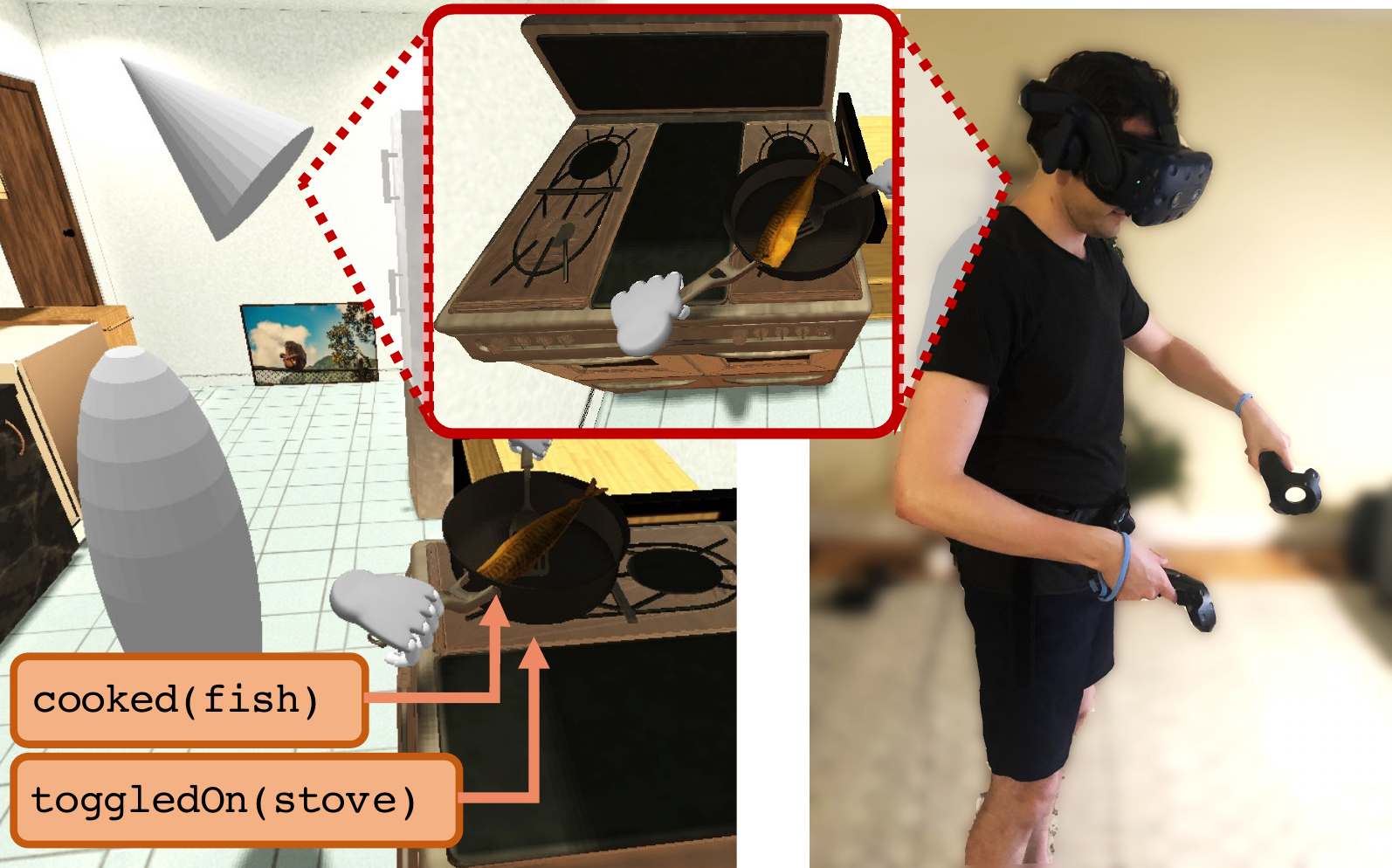}
    \caption{\textbf{Simulating new activities with \ig} (\textit{Left}) \ig's simulates a set of extended physical states (temperature, functional state, cleanliness, wetness level) for objects, enabling studying and developing solutions to new household tasks such as cooking. The full physical state can be mapped to symbolic representation that facilitates sampling new instances of tasks. (\textit{Right}) Humans can provide demonstrations for the new tasks with a novel virtual reality interface to enable policy learning. 
    }
    \label{fig:pullfig}
    \vspace{-1em}
\end{wrapfigure}

Third, to facilitate the development of new embodied AI solutions to new tasks in these new scenes, \ig includes a \textbf{new virtual reality interface} (VR) compatible with the two main commercially available VR systems. All states are logged during execution and can be replayed deterministically in the simulator, enabling the generation \textit{a posteriori} of additional virtual sensor signals or visualizations of the interactions and the development of imitation learning solutions. 

We evaluate the new functionalities of \ig on six novel tasks for embodied AI agents, and apply state-of-the-art robot learning algorithms to solve them. 
These tasks were not possible before in iGibson or in alternative simulation environments. 
Additionally, we evaluate the use of the new \ig VR interface to collect human demonstrations to train an imitation learning policy for bimanual operations. While the previous version of iGibson and other simulators provide interfaces to control an agent with a keyboard and/or a mouse, these interfaces are insufficient for bimanual manipulation.

In summary, \ig presents the following contributions:
\begin{itemize}[
    topsep=0pt,
    noitemsep,
    leftmargin=2pt,
    itemindent=12pt]
\item A set of new physical properties, e.g. temperature, wetness and cleanliness level, maintained and updated by the simulator; and a set of unary and binary logical predicates that map simulated states to a logical state that have a direct connection to semantics and language,
\item A set of generative functions associated with the logical predicates to sample valid simulated states from a given logical state, and a new rule-based mechanism exploiting these functions to populate the iGibson scenes with small objects placed at semantically meaningful locations to increase realism,
\item A novel virtual reality interface that allows humans to collect demonstrations for robot learning,
\end{itemize}
We hope that \ig will open new avenues of research and development in embodied AI, enabling solutions to household activities that have been under-explored before.

%% file: 2-rw.tex
\section{Related Work}
\label{s:rw}

\textbf{Simulation environments with (mostly) kinodynamic simulation:} In the last years, the robotics and AI communities have presented several impressive simulation environments and benchmarks: iGibson~\cite{xia2020interactive, shen2020igibson}, Habitat AI~\cite{habitat19arxiv}, AI2Thor (and variants)~\cite{kolve2017ai2,ehsani2021manipthor}, ThreeDWorld~\cite{gan2020threedworld}, Sapien~\cite{xiang2020sapien},  Robosuite~\cite{zhu2020robosuite}, VirtualHome~\cite{puig2018virtualhome}, RLBench~\cite{james2019rlbench}, MetaWorld~\cite{yu2020meta}, and more. They are based in physics engines such as (py)bullet~\cite{coumans2016pybullet}, MuJoCo~\cite{mujoco}, and Nvidia Physx~\cite{rieffel2009evolving, physx}, combined with rendering capabilities and usually enriched with a dataset of objects and/or scenes to use to develop embodied AI solutions. While these simulators have fueled research with new possibilities for training, testing and developing robotic solutions, they have skewed, with few exceptions, the exploration towards activities related to what they can simulate accurately: changes in kinematic states of (rigid or flexible) objects, i.e. Rearrangement tasks~\cite{batra2020rearrangement}. However, many everyday activities require the simulation of other physical states that can be modified by the agents, like the temperature of objects and their level of wetness or cleanliness. 
Recent simulators have attempted realistic simulation of fluids and flexible materials~\cite{rieffel2009evolving,flex,coevoet:hal-01649355,faure:hal-00681539} or extended rigid body simulators to approximate the dynamics of soft materials~\cite{coumans2016pybullet,lin2020softgym}.
Compared with these simulators, \ig provides a simple but effective mechanism to simulate the temperature of objects that change based on proximity to heat sources. It also simulates fluids through a system of droplets that can be absorbed by objects and change their wetness level.
Despite simpler than the accurate simulation of heat transfer, fluids and soft materials used by a few other simulators, \ig object-centric solution leads to realistic robotic behavior and motion in tasks involving changing temperature, handling liquids, or soaking objects.

\textbf{Simulation environments with object-centric representation:} In robotics, some simulators have adopted an object-centric representation with extended physical states, e.g. AI2Thor~\cite{kolve2017ai2} and VirtualHome~\cite{puig2018virtualhome}. Both are based on Unity~\cite{goldstone2009unity} and share a common set of functionalities. Actions in these simulators are predefined, discrete and symbolic, and characterized by preconditions (``what conditions need to be fulfilled for this action to be executable?'') and postconditions (``what conditions will change after the execution of this action?''), similar to actions in the planning domain definition language (PDDL~\cite{mcdermott1998pddl}) or STRIPS~\cite{fikes1971strips} but with the additional link to visual rendering of the predefined action execution and outcome.
In AI2Thor and VirtualHome, some of the actions change the temperature of an object between two or more discrete values pre- and post-execution of an action, e.g. \texttt{raw} and \texttt{cooked}.
This is fundamentally different to our approach in \ig: instead of maintaining only a symbolic state (\texttt{raw}/\texttt{cooked}), we provide a simple simulation of the underlying physical process (e.g. heat transfer) leading to continuously varying values of temperature and other extended states. These states are then \textit{mapped} into a symbolic representation through predicates (see Sec.~\ref{s:ls}). 
This provides a new level of detail in the execution of actions, where the agent can and should control the specific value of the object's extended states (temperature, wetness, cleanliness) to achieve a task, leading to more complex activities and more realistic execution. In the Appendix, we include a detailed comparison between \ig and other simulation environments in Table~\ref{t:simenv}.

\textbf{Simulation environments with virtual reality interfaces:} 
Researchers have used virtual reality interfaces before to develop robotic solutions with real robots~\cite{zhang2018deep, delpreto2020helping, robotlearningvr}.
VR has also been used in simulation environments to collect demos. VRKitchen~\cite{gao2019vrkitchen} collected demos for five cooking tasks in simulation. While realistic looking, activities in VRKitchen are performed with primitive actions similar to the pre-condition/post-condition system of AI2Thor and VirtualHome, falling short of realistic motion. More physically realistic are the VR interactions with UnrealROX/RobotriX~\cite{martinez2018unrealrox, garcia2018robotrix} and ThreeDWorld~\cite{gan2020threedworld}, based on Unreal~\cite{unreal} and Nvidia Physx~\cite{physx}. 
Our interface also enables realistic manipulation of objects, with additional features such as gaze tracking and assistive grasping to bridge the differences between simulation and the real-world.


%% file: 3a-extendedstates.tex
\section{Extended Physical States for Simulation of Everyday Household Tasks}
\label{s:extendedstates}

To perform household tasks, an agent needs to change objects' states 
beyond their poses. 
\ig extends objects with five additional states: temperature, $T$, wetness level, $w$, cleanliness level (dustiness level, $d$, or stain level, $s$), toggled state, $\textit{TS}$, and sliced state, $\textit{SS}$.
While some of these states could be different for different parts of an object, in \ig we simplify their simulation and adopt an \textbf{object-centric representation}: the simulator maintains a single value of each extended state for every simulated object (rigid, flexible, or articulated). This simplification is sufficient to simulate realistically household tasks such as cooking or cleaning.  We assume that the extended properties are \texttt{latent}: agents are not able to observe them directly. Therefore, \ig implements a mechanism to change objects' appearance based on their latent extended states (see Fig.~\ref{f:es1}), so that visually-guided agents can infer the latent states from sensor signals.

We further impose in \ig that every simulated object should be an instance of an existing object category in WordNet~\cite{miller1995wordnet}. This semantic structure allows us to associate characteristics to all instances of the same category~\cite{tversky1984objects,barsalou1999perceptual}. For example, we further simplify the simulation of extended states by annotating what extended states each category need. Not all object categories need all five extended states (e.g., the temperature is not necessary/relevant for non-food categories for most tasks of interest). The extended states required by each object category are determined by a crowdsourced annotation procedure in the WordNet hierarchy. 

In the following, we explain the details of the five extended states and the way they are updated in \ig. For a full list of object states (kinematics and extended), see Table~\ref{table:objectstates}.

\input{es1}

\textbf{Temperature:} 
To update temperature, \ig needs to approximate the dynamics of heat transfer from heat sources (hot) or sinks (cold). To that end, we annotate object categories in the WordNet hierarchy as \texttt{heat sources} or \texttt{heat sinks}. Heat sources elevate the objects' temperature over room temperature (23$^\circ$) towards the source's heating temperature, also annotated per category; similarly, heat sinks decrease the temperature under room temperature towards the sink's cooling temperature. The rate the objects change their temperature towards the temperature of the source/sink is a parameter annotated per category with units $^\circ$ C$/\SI{}{\second}$. We consider two types of heat source/sink: source/sinks that change the temperature of objects if they are proximal enough (e.g., a stove) and source/sinks that change the temperature of objects if they are inside of them (e.g., a fridge or an oven). 
Additionally, some of the sources/sinks need to be toggled on (see Functional State below) before they can change the temperature of other objects (e.g., a microwave).
For each object that can change temperature, the simulator first evaluates if it fulfills the conditions to be heated or cooled by a heat source/sink. Assuming that an object with temperature $T_o$ fulfills the conditions of a source/sink with heating/cooling temperature $T$ and changing rate $r$, the temperature of the object at each step is updated by $T_o^{t+1}=T_o^{t}\left(1 + \Delta_{sim} \times r \times (T - T_o^{t})\right)$, where $\Delta_{sim}$ is the simulated time. 

\ig maintains also a historical value of the \texttt{maximum temperature} that each object has reached in the past, $T_o^\text{max} = \max T_o^{t}$ for $t\in[0,\ldots,t_\text{now}]$. This value dictates the appearance of an object: if the object reached cooking or burning temperature in the past, it will look cooked or burned, even if its current temperature is low. 
Fig.~\ref{fig:temp} depicts the temperature system in action. 


\textbf{Wetness Level:} Similar to temperature, \ig maintains the level of wetness for each object that can get \texttt{soaked}. This level corresponds to the number of \textit{droplets} that have been absorbed by the object. 
In \ig, the system of droplets approximates liquid/fluid simulation. Specifically, droplets are small particles of liquid that are created in droplet sources (e.g. faucets), destroyed by droplet sinks (e.g. sinks), and absorbed by soakable objects (e.g. towels). They can also be contained in receptacles (e.g. cups) and poured later, leading to realistic behavior for the simulation of several household activities involving liquids, 
illustrated in Fig.~\ref{fig:wetness}.

\input{es2}

\textbf{Cleanliness -- Dustiness and Stain Level:} A common task for robots in homes and offices is to clean dirt. This dirt commonly appears in the form of dust or stains. In \ig, the main difference between dust and stains is the way they get cleaned: while dust can be cleaned with a dry cleaning tool like a cleaning cloth, stains can only be cleaned with a soaked cleaning tool like a scrubber. To clean a particle of dirt (dust or stain), the right part of a cleaning tool 
should get in physical contact with the particle. Once a dirt particle is cleaned, it disappears from objects' surface.

In \ig, objects can be initialized with visible dust or stain particles on its surface. The number of particles at initialization corresponds to a  100\% level of dustiness, $d$, or stains, $s$, as we assume that dust/stain particles cannot be generated after initialization. As particles are cleaned, the level decreases proportionally to the number of particles removed, reaching a level of 0\% dustiness, $d$, or stains when the object is completely clean (no particles left). This extended state allows simulating multiple cleaning tasks in our simulator: the agent needs to exhibit a behavior (motion, use of tools) similar to the one necessary in the real world. Fig.~\ref{fig:cleanli} depicts an example of the cleanliness level simulation in \ig, for both dust and stain particles.

\textbf{Toggled State:} Some object categories in \ig can be toggled on and off. \ig maintains and updates an internal binary functional state for those objects. The functional state can affect the appearance of an object, but also activate/deactivate other processes, e.g. heating food inside a microwave requires the microwave to be toggled on. To toggle the object, a certain area needs to be touched. For object models of categories that can be toggled on/off, we annotate a \texttt{TogglingLink}, an additional virtual fixed link that needs to be touched by the agent to change the toggled state. Fig.~\ref{fig:functionalstate} depicts an example of an object that can change its toggled state, an oven. 

\textbf{Sliced State:} Many cooking activities require the agent to slice objects, e.g. food items. Slicing is challenging in simulation environments where objects are assumed to have a fixed (rigid or flexible) 3D structure of vertices and faces. To approximate the effect of slicing, \ig maintains and updates a sliced state in instances of object categories that are annotated as \texttt{sliceable}. When the sliced state transitions to True, the simulator replaces the whole object with two halves. The two halves will be placed at the same location and inherit the extended states from the whole object (e.g. temperature). The transition is not reversible: the object will remain sliced for all upcoming simulated time steps. Objects can only be sliced into two halves, with no further division. The sliced state changes when the object is contacted with enough force (over a slicing force threshold for the object) by a slicing tool, e.g. a knife. Objects of these categories are annotated as \texttt{SlicingTool}. If an object is a slicing tool, it will undergo a second annotation process to obtain a new virtual fixed link that acts as \texttt{SlicingLink}, the part of the slicing tool that can slice an object, e.g., the sharp edge of a knife.
Fig.~\ref{fig:slicedstate} depicts an example of a peach being sliced by a knife. For more information about update rules for all extended physical states, please refer to see Sec.~\ref{ss:kinematicsampling}.

%% file: es1.tex
\begin{figure}[t]
\begin{subfigure}[t!]{0.99\linewidth}
\centering
\includegraphics[width=0.19\linewidth]{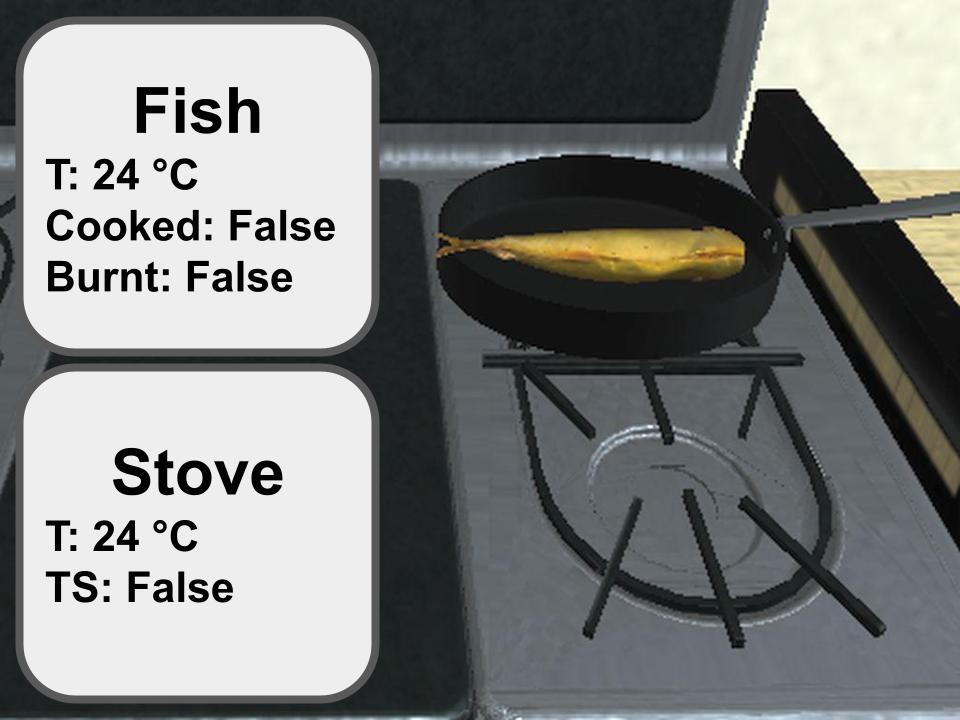}
\hfill
\includegraphics[width=0.19\linewidth]{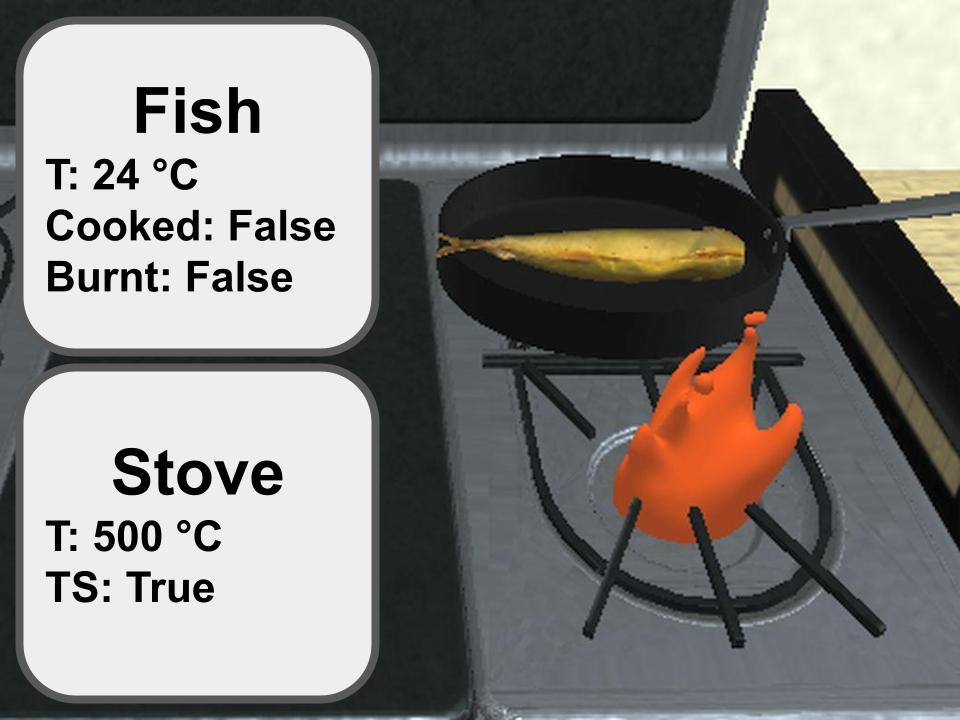}
\hfill
\includegraphics[width=0.19\linewidth]{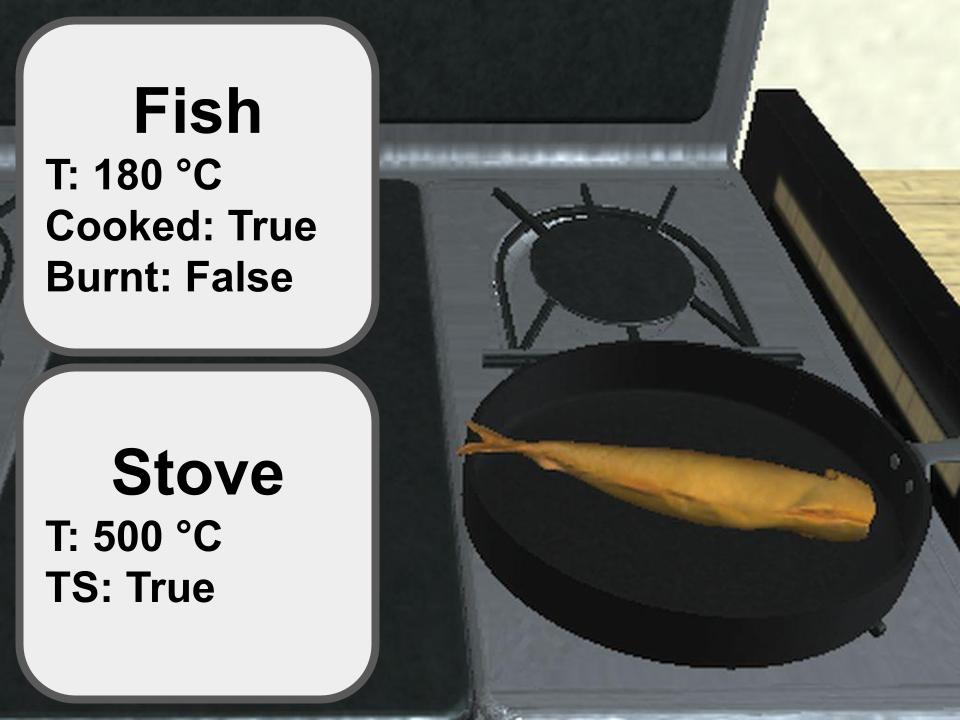}
\hfill
\includegraphics[width=0.19\linewidth]{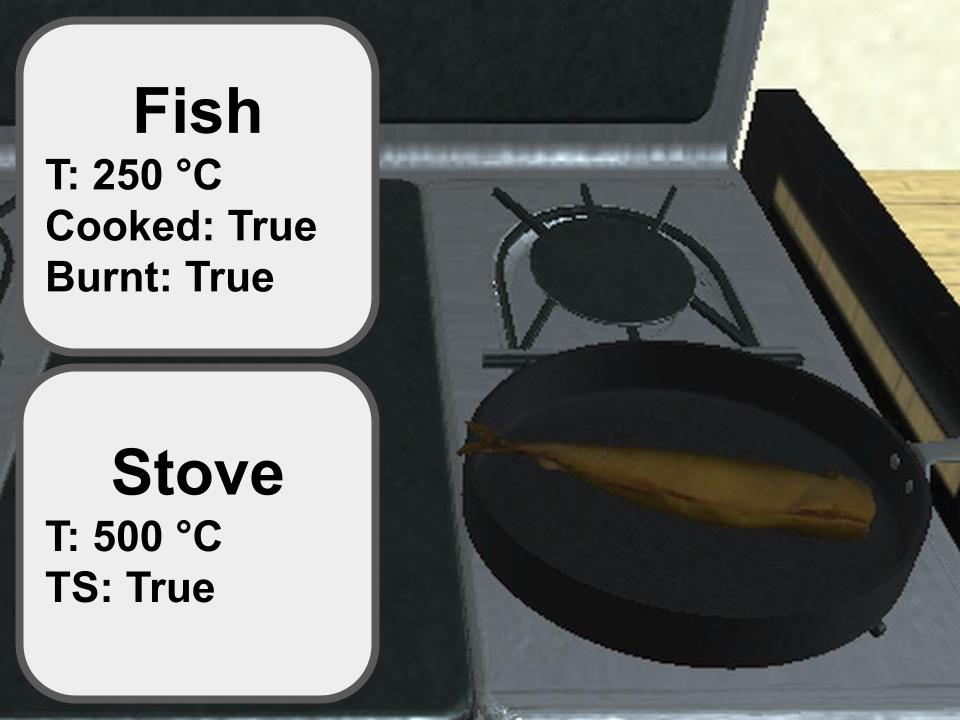}
\hfill
\includegraphics[width=0.19\linewidth]{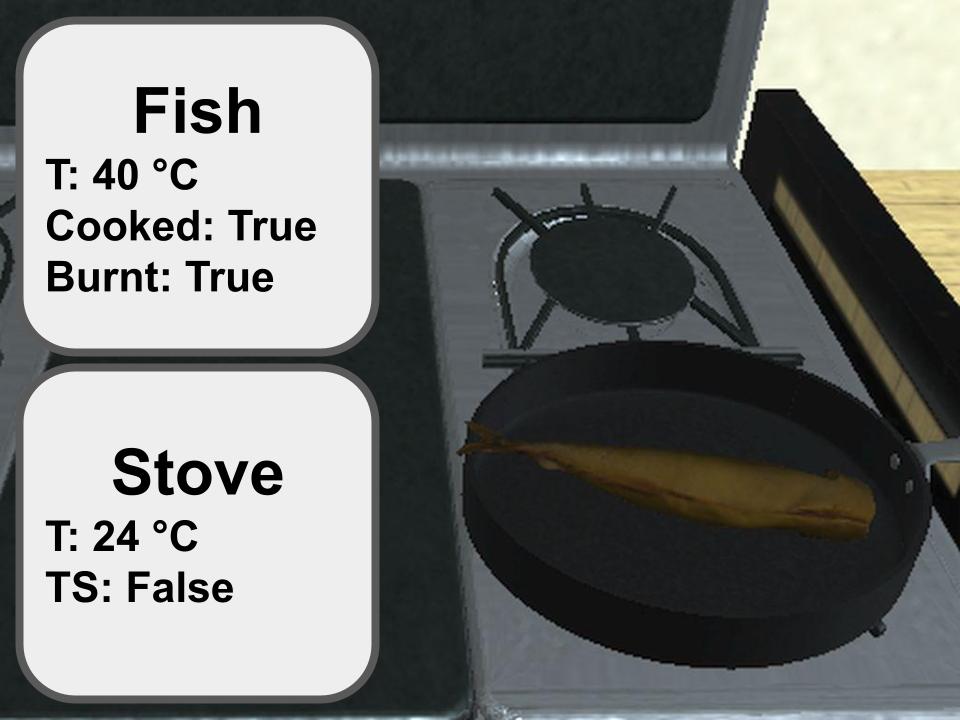}
\caption{\textbf{Object temperature:} \textit{(From left to right)} A stove --toggleable heat source by proximity-- is toggled on (second from left), and starts to heat a nearby fish. The temperature and appearance of the fish changes due to proximity to the heat source reaching the temperature to be cooked (third from left). Additional heat elevates further the fish's temperature, burning it (fourth from left). After the stove is toggled off, the fish changes back to room temperature, keeping the appearance that corresponds to the maximum temperature reached (most right). The temperature system with heat sources and sinks, enable visually guided execution of household activities such as cooking.
}
\label{fig:temp}
\end{subfigure}
\begin{subfigure}[t!]{0.99\linewidth}
\centering
\includegraphics[width=0.24\linewidth]{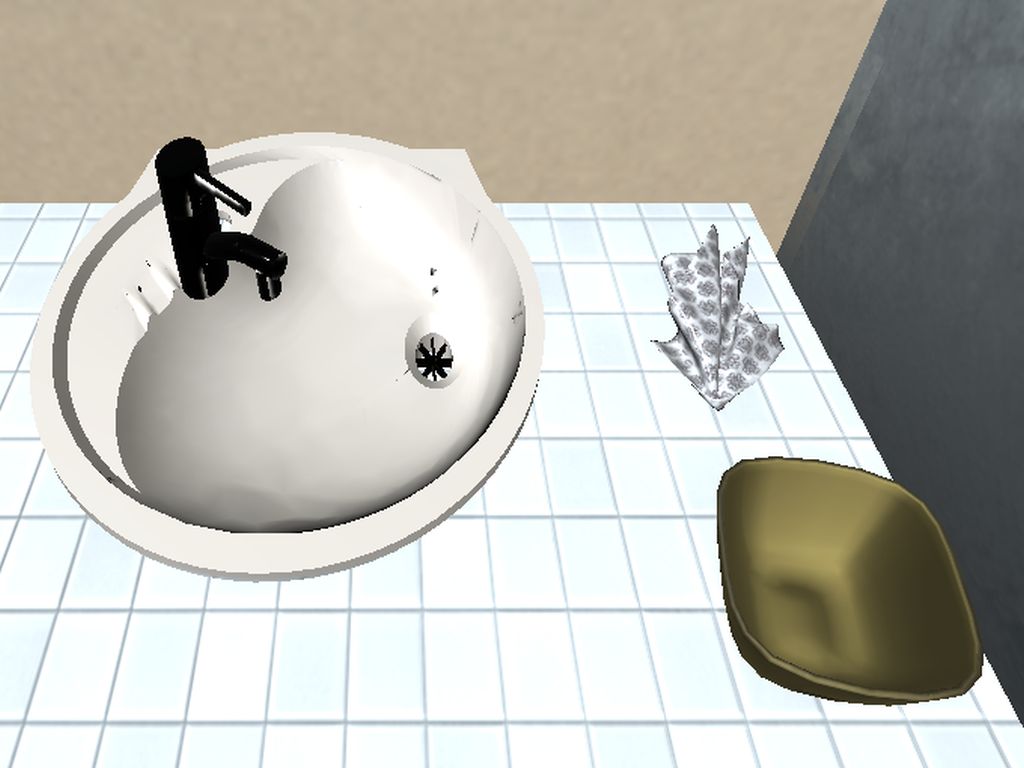}
\hfill
\includegraphics[width=0.24\linewidth]{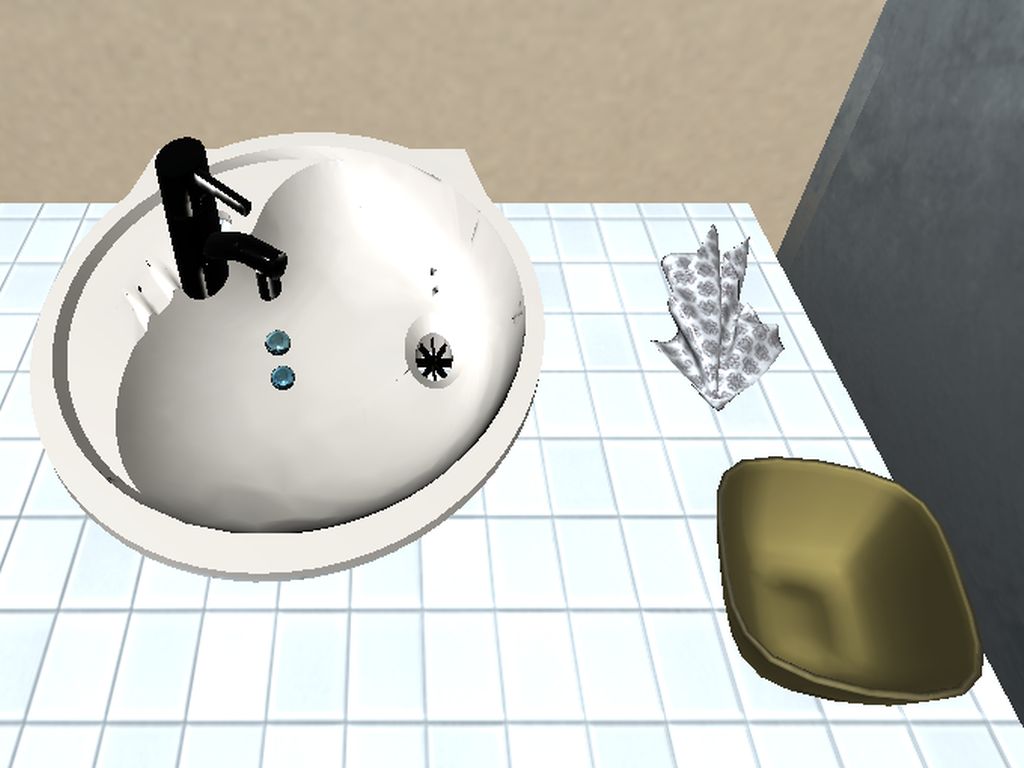}
\hfill
\includegraphics[width=0.24\linewidth]{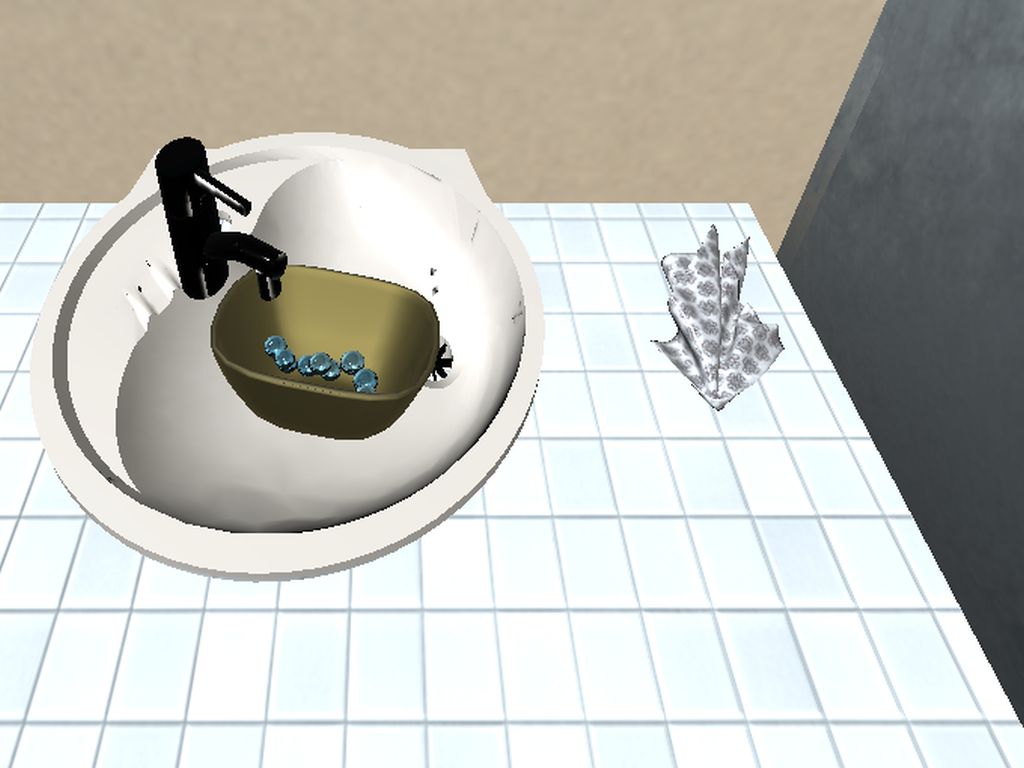}
\hfill
\includegraphics[width=0.24\linewidth]{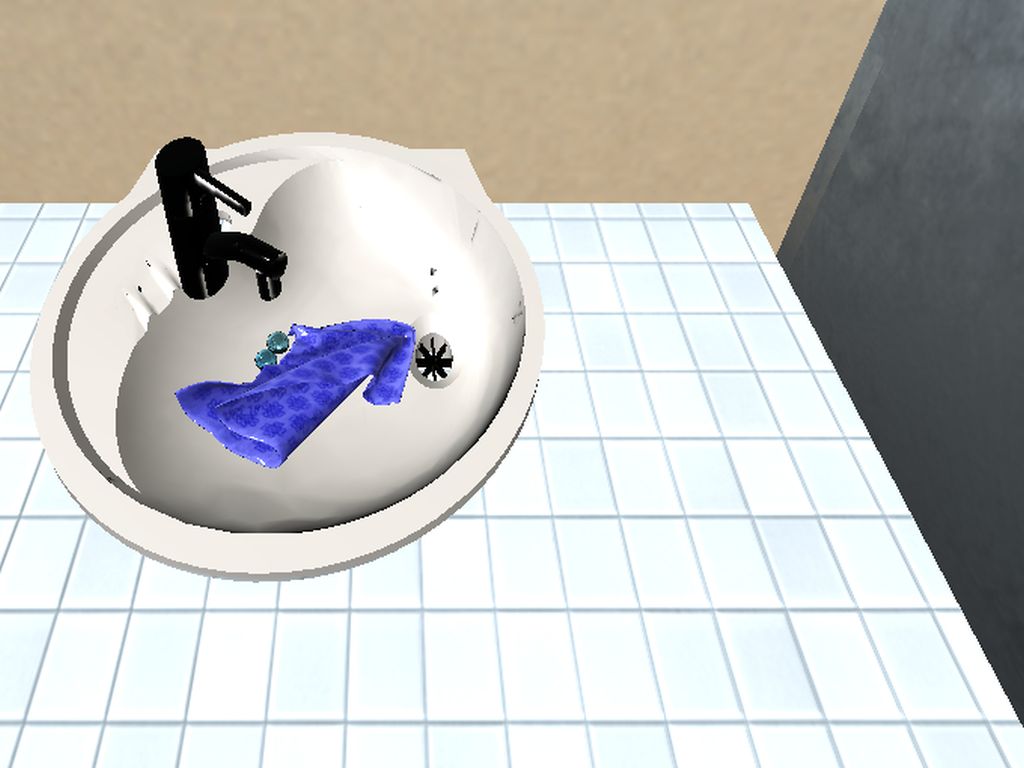}
\caption{\textbf{Object wetness level:} \textit{(From left to right)} A sink  --toggleable droplet source-- is toggled on, and starts to create droplets (second from left). Droplets can be contained in a receptacle to be poured on other objects (third from left). An object that can change its wetness level gets in contact with the droplets and absorbs them, changing its appearance (most right). With the droplets system, iGibson v2.0 provides a simplified liquid simulation sufficient to perform common household activities.
}
\label{fig:wetness}
\end{subfigure}
\caption{Extended object states I): temperature and wetness level}
\label{f:es1}
\vspace{-8mm}
\end{figure}

%% file: es2.tex
\begin{figure}
\begin{subfigure}[t!]{0.99\linewidth}
\centering
\includegraphics[width=0.24\linewidth]{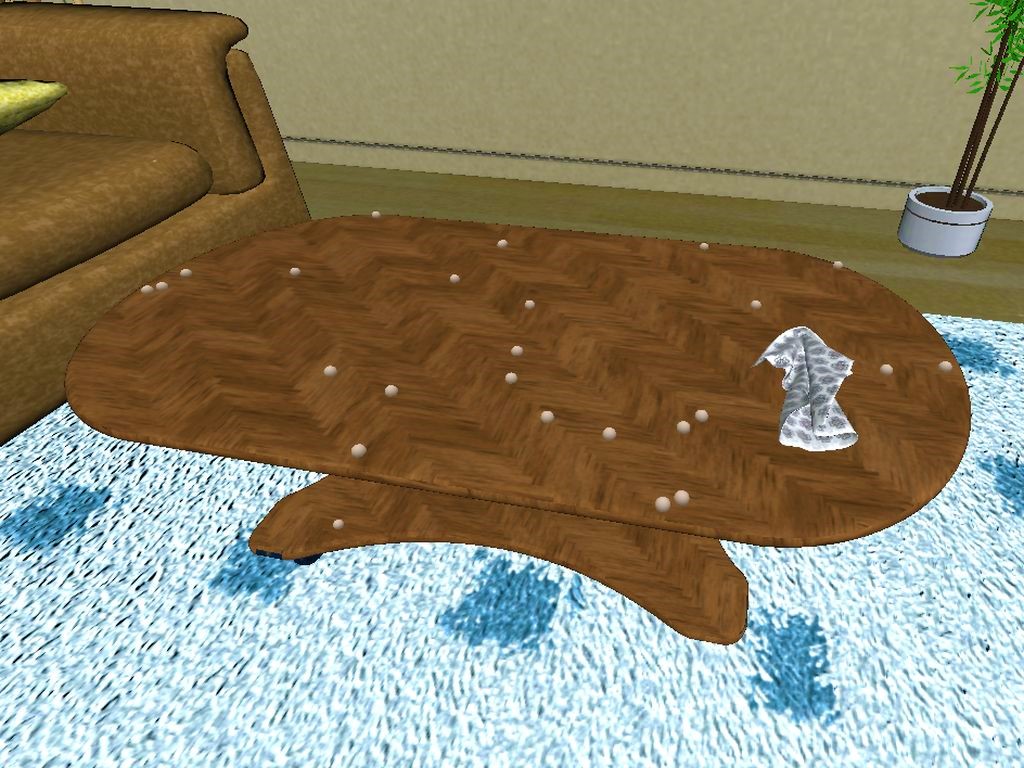}
\hfill
\includegraphics[width=0.24\linewidth]{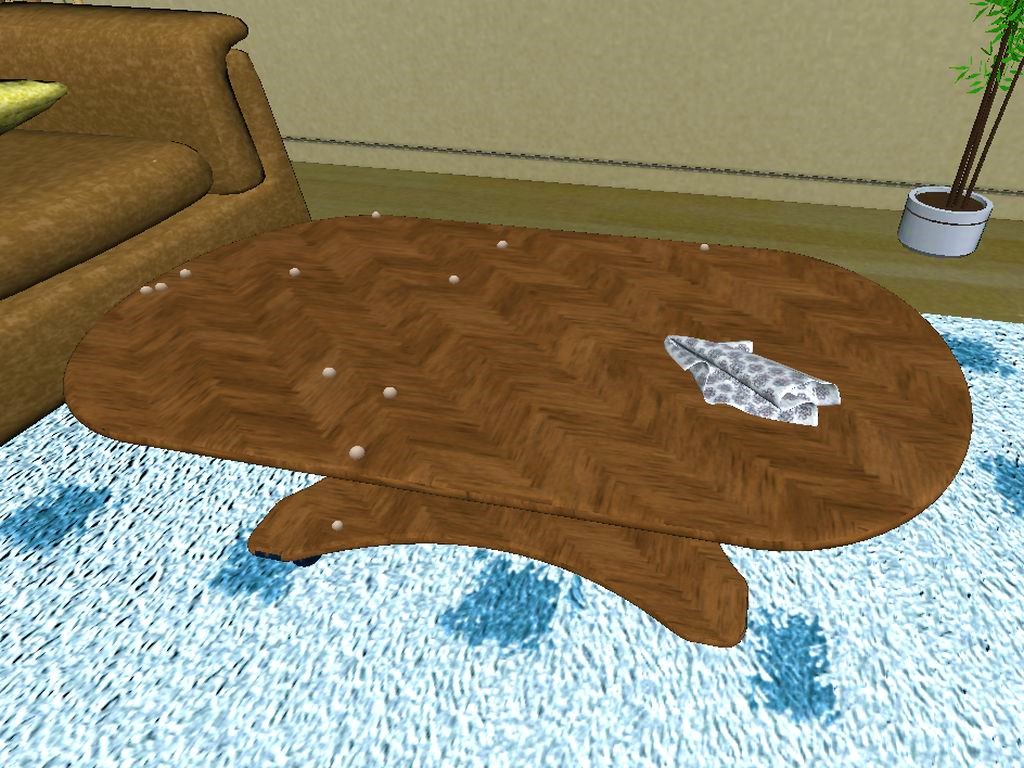}
\hfill
\includegraphics[width=0.24\linewidth]{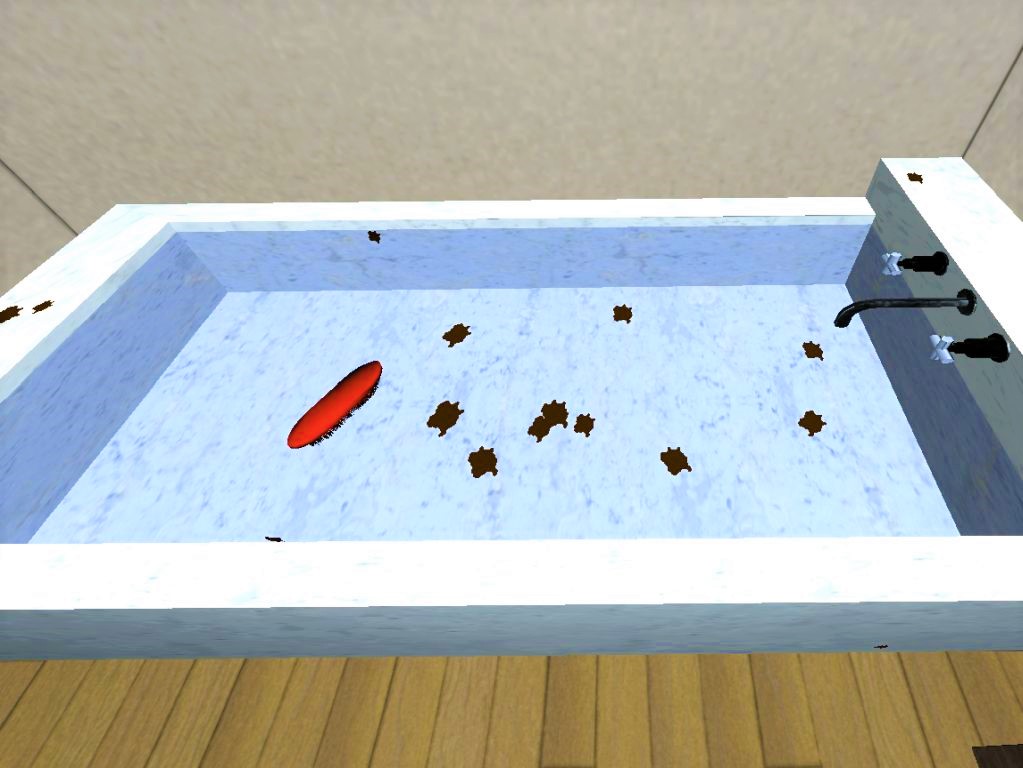}
\hfill
\includegraphics[width=0.24\linewidth]{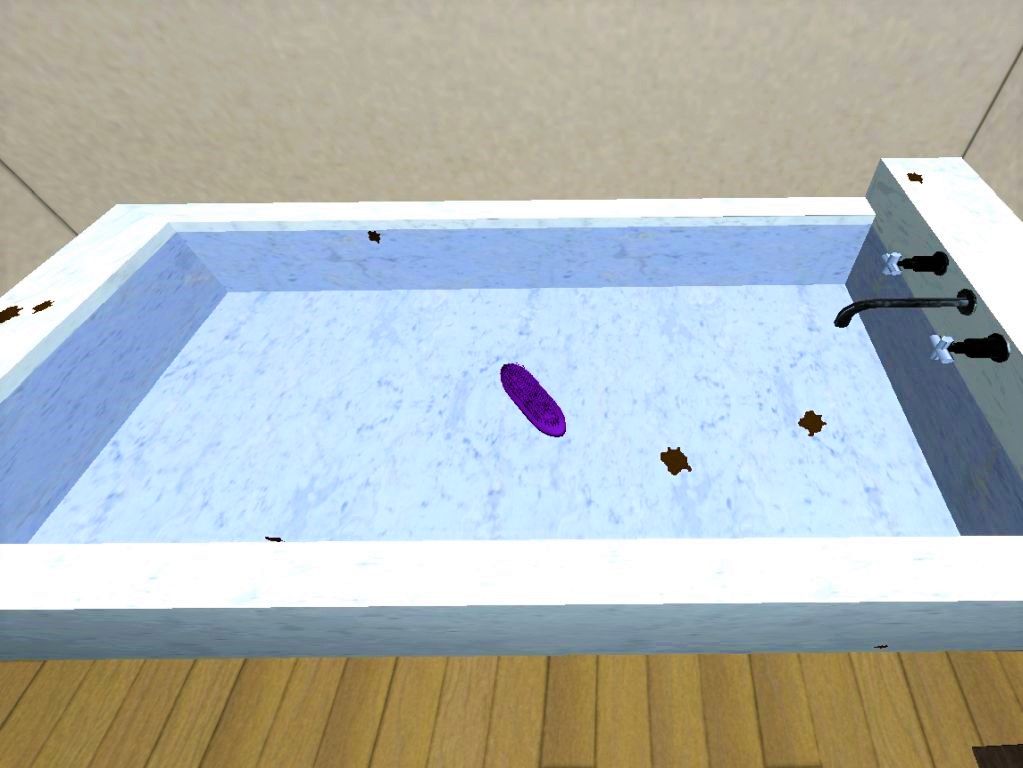}
\caption{\textbf{Object cleanliness level:} \textit{(From left to right)} An object is initialized with dust particles that can be cleaned with a cloth; Another object is initialized with stains that require a wet tool (a scrubber) to be removed; Realistic behaviors and motion are required in iGibson v2.0 to change the cleanliness level of the objects.
}
\label{fig:cleanli}
\end{subfigure}
\centering
\begin{subfigure}[t]{0.48\linewidth}%
\includegraphics[width=0.49\linewidth]{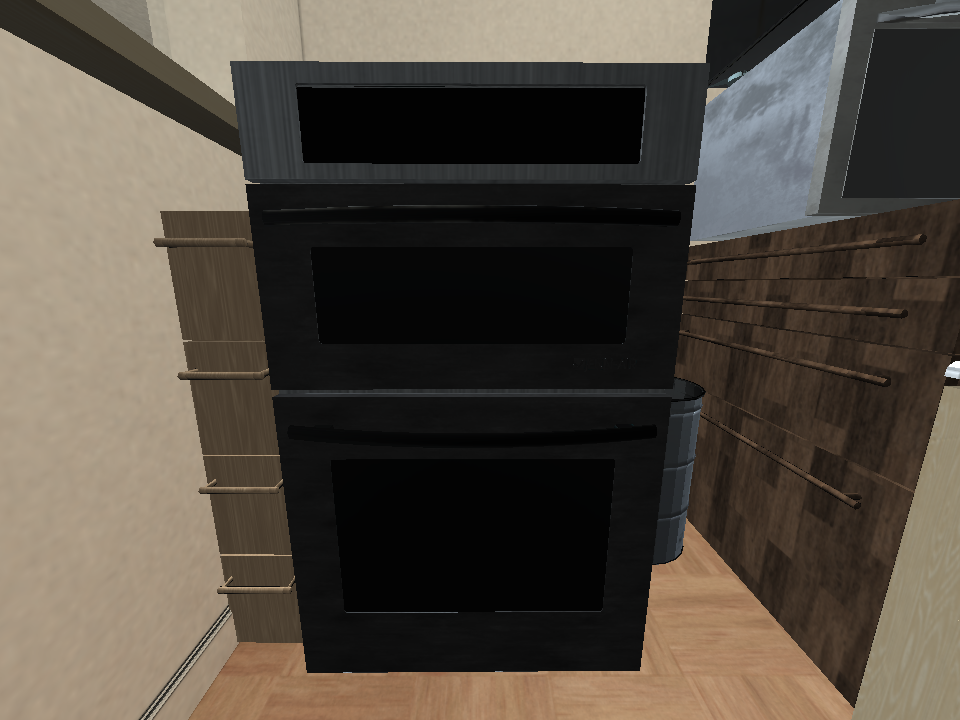}%
\hfill%
\includegraphics[width=0.49\linewidth]{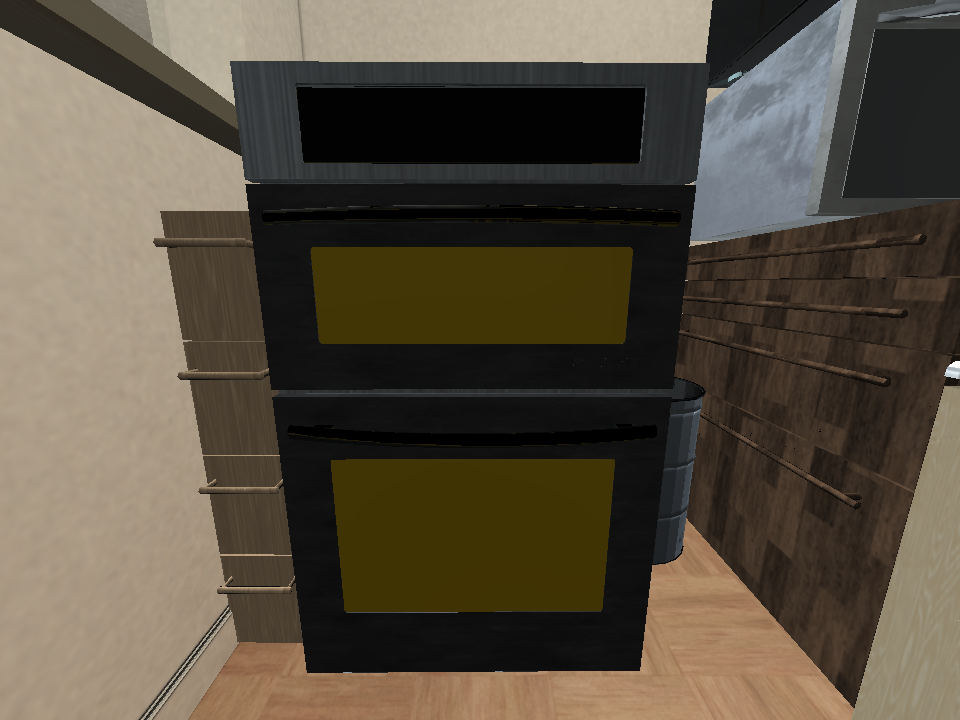}%
\caption{\textbf{Object toggling state:} An object with the toggled state, an oven, is initially off. When it is toggled on, its appearance changes indicating the transition.
}
\label{fig:functionalstate}
\end{subfigure}
\hfill
\begin{subfigure}[t]{0.48\linewidth}%
\includegraphics[width=0.49\linewidth]{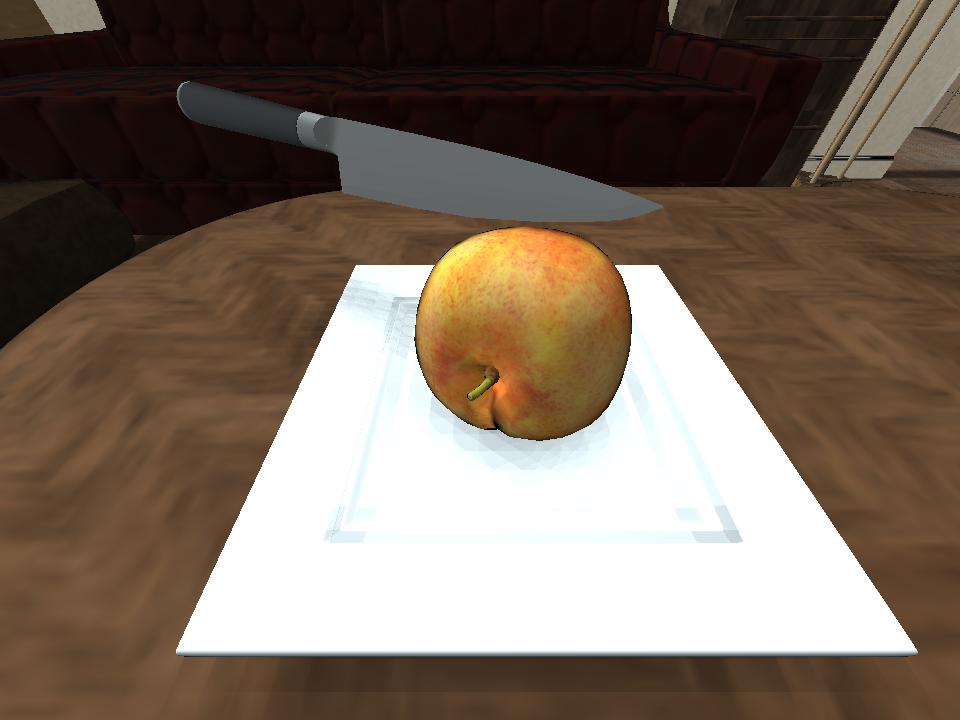}%
\hfill%
\includegraphics[width=0.49\linewidth]{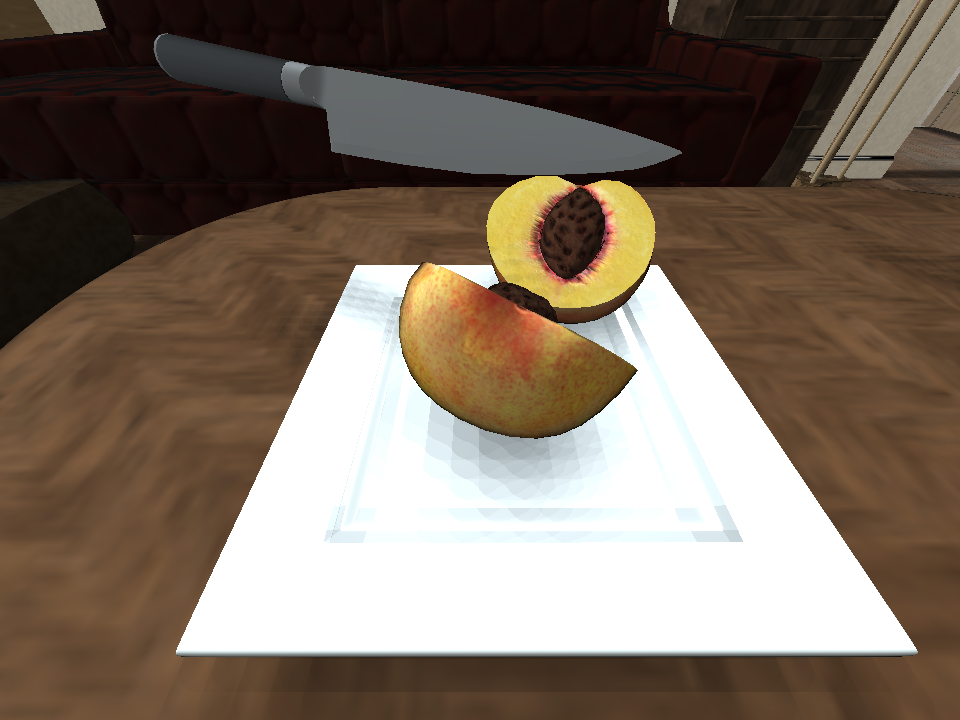}%
\caption{\textbf{Object slice state:} An object with the sliced state, a peach, is sliced after exerting enough force with the right part of a slicing tool (the sharp edge of a knife)
}
\label{fig:slicedstate}
\end{subfigure}
\caption{Extended object states II): cleanliness level (dustiness, stains), toggling, slice state}
\label{f:es2}
\vspace{-6mm}
\end{figure}

%% file: 3b-predicates.tex
\section{Logical Representation of Physical States}
\label{s:ls}

The new extended object states from \ig are sufficient to simulate a new set of household activities in indoor environments. However, there is a semantic gap between the extended states (e.g. temperature or wetness level) and the natural description of activities in a household setup (e.g. cooking apples).
To bridge this gap, 
we define a set of functions that map the extended object states to logical states for single objects and pairs of objects. The logical states are semantically grounded on common natural language representing properties such as \texttt{cooked} or \texttt{dusty}.

The list of logical predicates covers kinematic states between pair of objects (\texttt{InsideOf}, \texttt{OnTopOf}, \texttt{NextTo}, \texttt{InContactWith}, \texttt{Under}, \texttt{OnFloor}), states related to the internal degrees of freedom of articulated objects (\texttt{Open}), states based on the object temperature (\texttt{Cooked}, \texttt{Burnt}, \texttt{Frozen}), wetness and cleanliness level (\texttt{Soaked}, \texttt{Dusty}, \texttt{Stained}), and functional state (\texttt{ToggledOn}, \texttt{Sliced}). A complete list of the logic predicates with detailed explanation is included in Table~\ref{table:supp_obj_states}. They allow \ig to map a physical simulated state into an corresponding logical state.


\subsection{Generative System based on Logical Predicates}
\label{ss:gmlp}

Logical predicates map multiple physically simulated states to the same logical state, e.g., all relative poses between to objects that correspond to being \texttt{onTop}. In addition to this discriminative role, we include functionalities in \ig to use logical predicates in a generative manner, to describe initial states symbolically that can be used to initialize the simulator. \ig includes a sampling mechanism to create valid instances of tasks described with logical predicates. This mechanism facilitates the creation of multiple  semantically meaningful initial states, without the laborious process of manually annotating the initial distributions per scene.


The process of sampling valid object states is different depending on the nature of the logical predicate. For predicates based on objects' extended states such as \texttt{Cooked}, \texttt{Frozen} or \texttt{ToggledOn}, we just sample values of the extended states that satisfy the predicate's requirements, e.g., a temperature below the annotated freezing point of an object to fullfil the predicate \texttt{Frozen}. Particles for \texttt{Dusty} and \texttt{Stained} are sampled on the surface of an object following a pseudo-random procedure. Generating initial states to fulfill kinematic predicates such as \texttt{OnTopOf} or \texttt{Inside} is a more complex procedure as the underlying physical state (the object pose) must lead to a stationary state (e.g., not falling) that does not cause penetration between objects. Each kinematic predicate is implemented differently, combining mechanisms that include ray-casting and analytical methods to verify the validity of sampled poses. For example, to sample a state that satisfies \texttt{Inside(A, B)}, we implement a procedure that generates 6D poses inside of object \texttt{B} and we evaluate that 1) a bounding box of the the size of object \texttt{A} does not penetrate object \texttt{B} and rays cast from \texttt{A} intersect object \texttt{B} from  evaluated by casting rays from the the internal box. 
For more details on the sampling mechanism, see Sec.~\ref{ss:kinematicsampling}. 


\subsection{\ig Scenes with Realistic Object Distribution Created by Generative System}

\begin{wrapfigure}{r}{0.55\textwidth}
\vspace{-1em}
    \centering
    \includegraphics[width=0.45\linewidth]{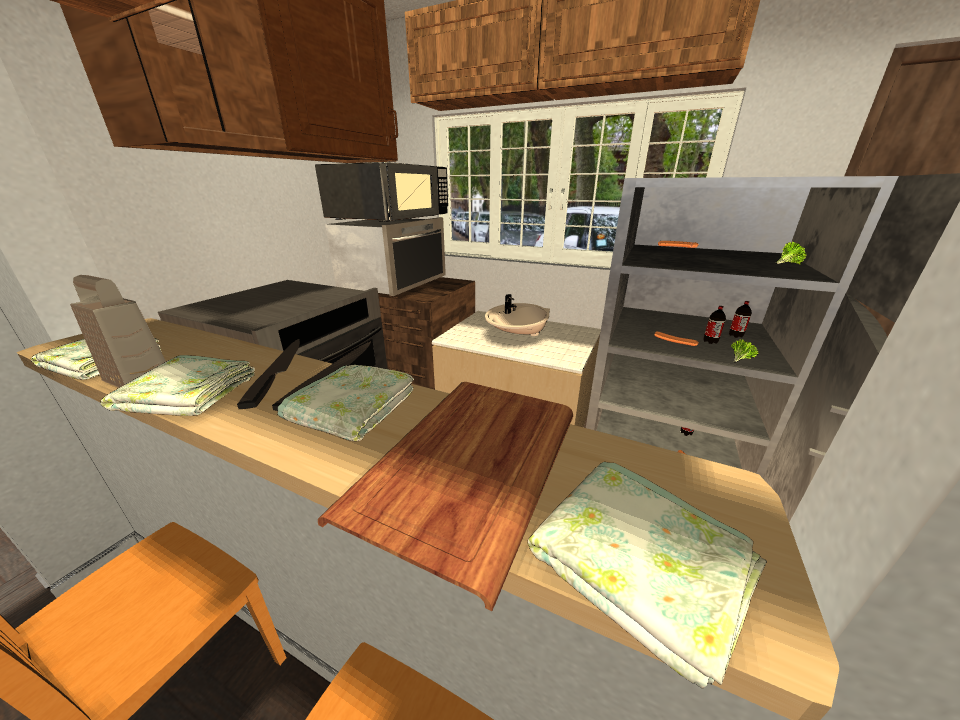}
    \hspace{0.2cm}
    \includegraphics[width=0.45\linewidth]{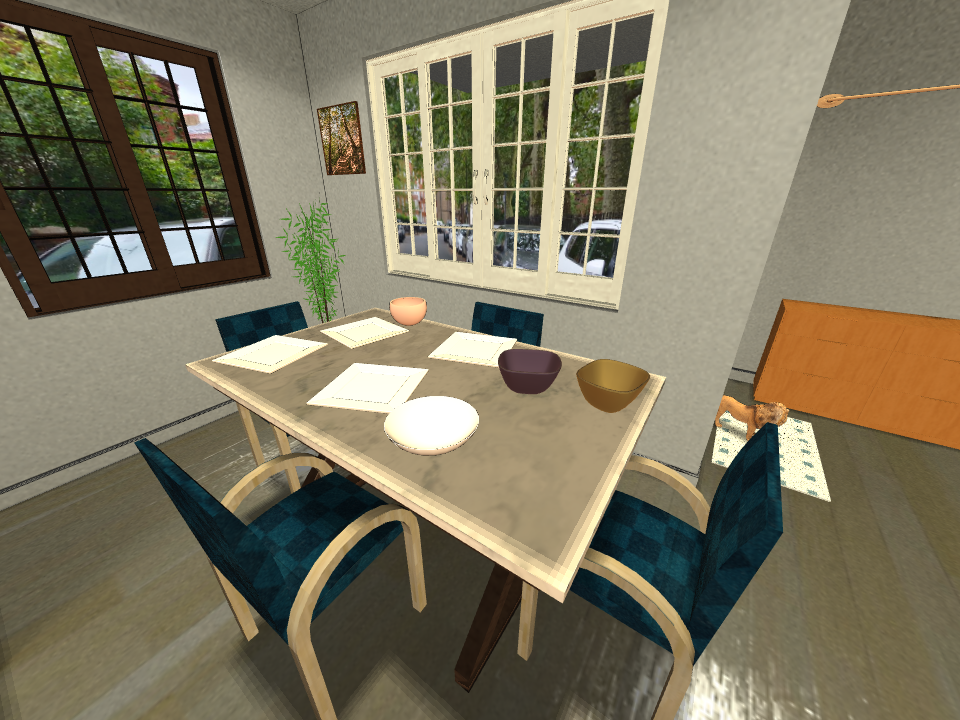}
    \caption{\textbf{Densely populated ecological scenes from the \ig dataset:} Objects are sampled in semantically meaningful locations using a set of semantic rules and the generative system based on logical statements, increasing realism for robot learning and VR experience.
    }
    \label{fig:populatedscenes}
    \vspace{-1em}
\end{wrapfigure}
One common issue that limits the realism of indoor scenes in simulation is that they are less densely populated than those in the real world.
Creating highly populated simulated houses is usually a laborious process that requires manually selecting and placing models of small objects in different locations. 
Thanks to the generative system in \ig, this process can be extremely simplified. The users only need to specify a list of logical predicates that represent a realistic distribution of objects in a house.

We provide as part of \ig a set of semantic rules to generate more populated scenes, and a new version of the iGibson original 15 fully interactive scenes, populated with additional small objects as a result of the application of the rules.
Given an indoor scene with multiple types of rooms (\texttt{kitchen}, \texttt{bathroom}, \ldots) that are populated with furniture containers and appliances (\texttt{fridge}, \texttt{cabinet}, \ldots), the semantic rules define the probabilities for object instances of diverse categories to be sampled in certain container type in a given room type, e.g. $p($\texttt{InsideOf(Beer, Fridge)}$\land$\texttt{InsideOf(Fridge, Kitchen)}$)$.
To generate the more densely populated versions of the 15 scenes, we first collect a large number of small 3D object models created by artists and we annotate them with semantic categories (e.g. \texttt{cereal}, \texttt{apple}, \texttt{bowl}) and realistic dimensions. Then, we apply in a random sequence the logic predicates using the generative system explained in Sec.~\ref{ss:gmlp}, increasing the number of objects in the scene by over 100. The result is depicted in Fig.~\ref{fig:populatedscenes}. 

%% file: 3c-VR.tex
\section{Virtual Reality Interface}

\label{s:vr}


\ig's new functionalities enable modeling new household activities and generating multiple instances in more densely populated scenes. 
To facilitate research in these new, complex tasks, \ig includes a novel virtual reality (VR) interface compatible with major commercially available VR headset through OpenVR~\cite{openvr}. One of the goals is to allow researchers to collect human demonstrations, and use them to develop new solutions via imitation (see Sec.~\ref{s:es}).

\ig's VR interface creates an immersive experience: humans embody an avatar in the same scene and for the same task as the AI agents. 
The virtual reality avatar (see Fig.~\ref{fig:pullfig}) is composed of a main body, two hands and a head. The human controls the motion of the head and the two hands via the VR headset and hand controllers with an optional, additional tracker for control of the main body. Humans receive stereo images as generated from the point of view of the head of the virtual avatar, at at least 30 fps (up to 90 fps) using the PBR rendering functionalities of iGibson~\cite{shen2020igibson}.

\textbf{Grasping in VR:} Grasping in the real-world, while natural to adult humans, is a complex experience that proves difficult to reproduce with virtual reality controllers. Our empirical observations revealed that while using solely physical grasping, user dexterity was significantly impaired relative to the real-world resulting in unnatural behavior when manipulating in VR. 
To provide a more natural grasping experience, we implement an assistive grasp (AG) mechanism that enables an additional constraint between the palm and a target object after the user passes a grasp threshold (50\% actuation) and provided the object is in contact with the hand, between the fingers and the palm. This facilitates grasping of small objects, and prevents object slippage. To not render grasping artificially trivial, the AG connection can break if the constraint is violated beyond a set threshold, such as while lifting heavy objects or during intense acceleration, encouraging natural task execution that leverages careful motions and bimanual manipulation. Please refer to Sec.~\ref{ss:vrdetails} for additional details about AG.

\textbf{Navigating in VR:} Navigation of the avatar is controlled by the locomotion of the human. However, the VR space is much smaller than the typical size of \ig scenes. To navigate between rooms, we configured a touchpad in the hand controller that humans can use to translate the avatar.


%% file: 4-eval.tex
\vspace{-1mm}
\section{Evaluation}
\vspace{-1mm}
\label{s:eval}

\begin{figure}[t]
\begin{subfigure}[t!]{0.6\linewidth}
\centering
\includegraphics[width=0.3\linewidth]{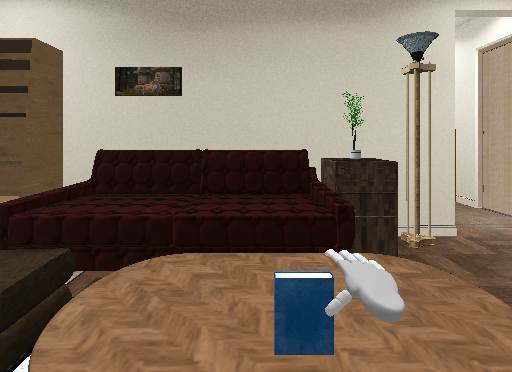}
\hfill
\includegraphics[width=0.3\linewidth]{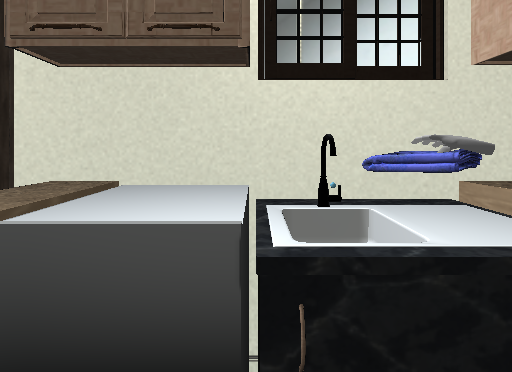}
\hfill
\includegraphics[width=0.3\linewidth]{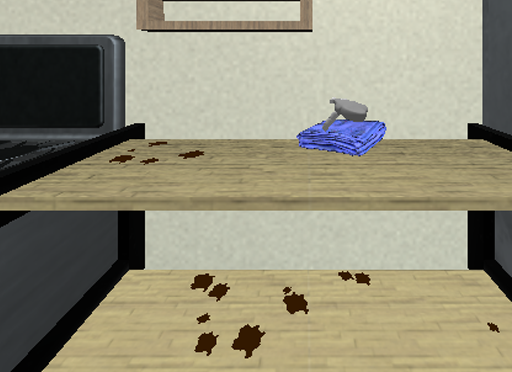}
\\\vspace{0.1cm}
\includegraphics[width=0.3\linewidth]{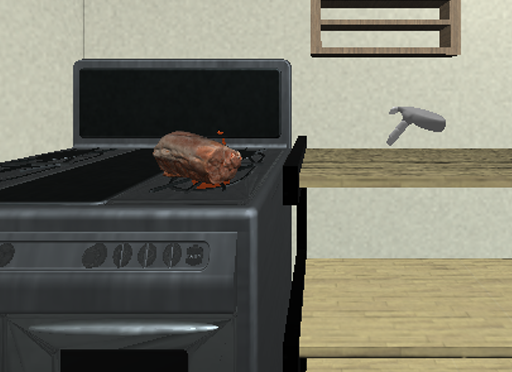}
\hfill
\includegraphics[width=0.3\linewidth]{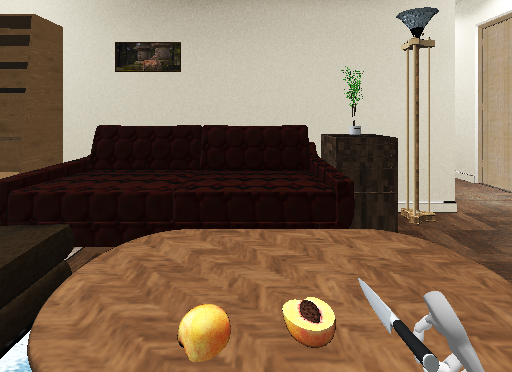}
\hfill
\includegraphics[width=0.3\linewidth]{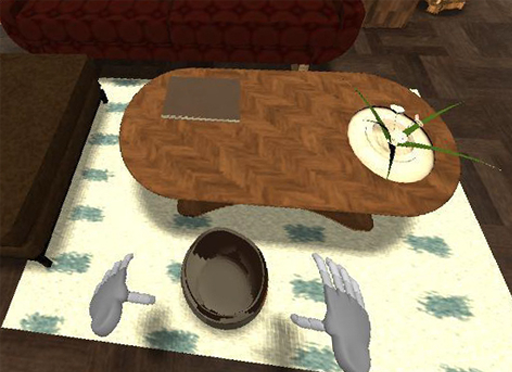}
\caption{\textbf{Six evaluated tasks}: (top) 1) \textit{Grasping Book}, 2) \textit{Soaking Towel}, 3) \textit{Cleaning Stained Shelf}; (bottom) 4) \textit{Cooking Meat}, 5) \textit{Slicing fruit}, 6) \textit{Bimanual Pick and Place}.
}
\label{fig:exptasks}
\end{subfigure}
~
\begin{subfigure}[t!]{0.34\linewidth}

\centering
\includegraphics[width=\linewidth]{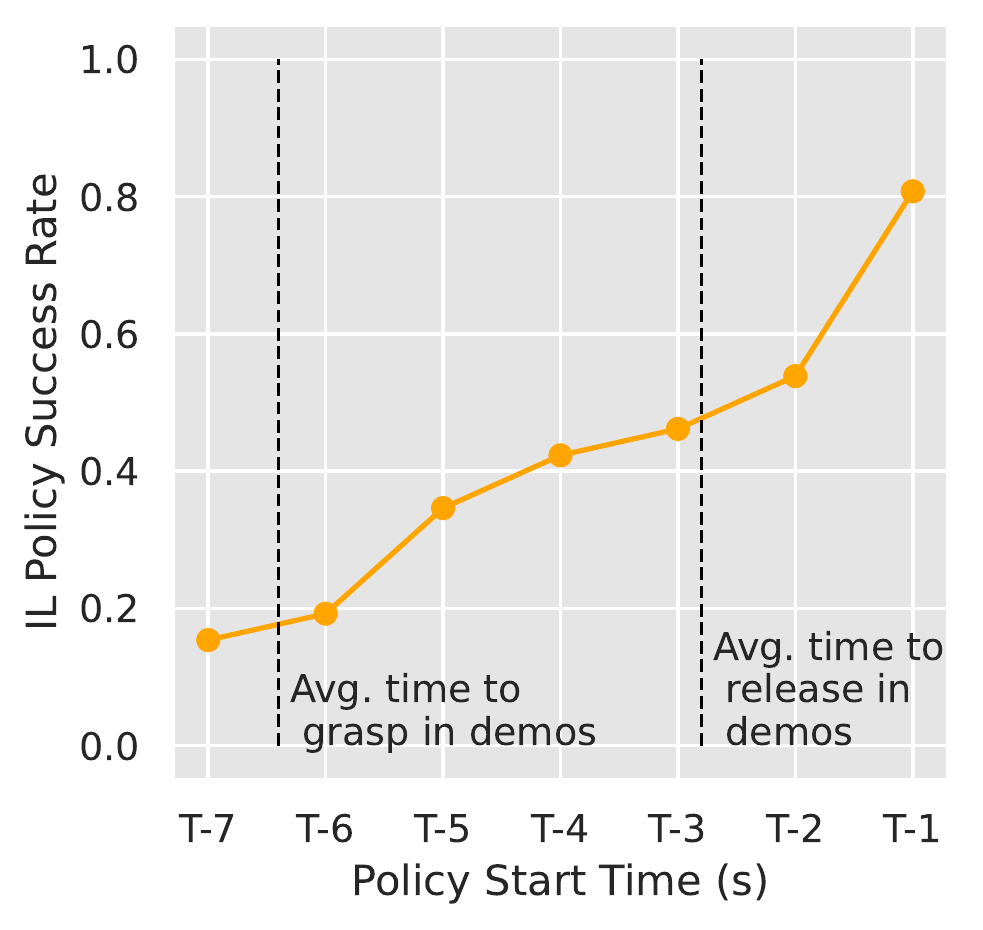}
\vspace{-2em}
\caption{\textbf{Success rate of IL in  \textit{Bimanual Pick and Place}}}
\label{fig:expcurves}
\end{subfigure}
\caption{\textbf{Evaluation:} We create six novel household tasks in \ig where the agents have to manipulate extended states. These tasks cannot be studied in other simulation environments.}
\label{f:expres}
\vspace{-4mm}
\end{figure}

In our evaluation, we test the new functionalities of \ig explained above and that sets it apart from other existing environments. First, we create a set of tasks that showcase the new extended states (see Fig.~\ref{fig:exptasks}) as their modification is required to achieve the tasks. In our experiments, we make use of our discriminative and generative logical engine to detect task completion and create multiple instances of each task for training. Then, we use our novel VR interface to collect human demonstrations to train an imitation learning policy for a bimanual task.

\vspace{-8pt}\paragraph{Experimental Setup:}
\label{s:es}
To evaluate and demonstrate the new capabilities of \ig, we create the following six novel tasks for embodied AI agents: \textbf{1)} \textit{Grasping Book:} the agent has to grasp a book \texttt{OnTopOf} a table and lift it. This task demonstrates the kinematic logical states; \textbf{2)} \textit{Soaking Towel:} the agent has to soak a cleaning tool (a towel) with water droplets from a sink. This task demonstrates the liquid/droplets system. \textbf{3)} \textit{Cleaning Stained Shelf:} the agent has to clean a stained shelf using a cleaning tool. This task showcases the capabilities to update the cleanliness level of an object. \textbf{4)} \textit{Cooking Meat:} the agent has to cook a piece of meat by placing it on a heat source (a burning stove) and waiting for enough time for the temperature to rise. This task showcases the capabilities to update the temperature of an object. \textbf{5)} \textit{Slicing Fruit:} the agent has to slice a fruit by exerting enough force with the \texttt{SlicingLink} of a slicing tool (a knife). This task showcases the capabilities to simulate sliceable objects and to model the interaction between cutting tools and sliceable objects. \textbf{6)} \textit{Bimanual Pick and Place:} The agent has to pick up a heavy object (a cauldron) and place it \texttt{OnTopOf} a table. Manipulating this object requires bimanual interaction as its mass is $\SI{50}{\kilo\gram}$ and each hand can only exert at most $\SI{300}{N}$. Demonstrating bimanual manipulation is natural and easy with our novel VR interface, which is not the case with previous interfaces such as keyboard~\cite{shen2020igibson}.



We conduct two sets of experiments to evaluate the current robot learning algorithms on these tasks. First, we train agents with a state-of-the-art reinforcement learning (RL) algorithm, Soft-Actor Critic (SAC~\cite{haarnoja2018soft}). For embodiment, we use a bimanual humanoid robot (the one used in VR) and a Fetch robot. The agent receives RGB-D images from its onboard sensors and proprioception information as observation and outputs the desired linear and angular velocities of the right hand (assuming the rest of the agent is stationary). We adopt the ``sticky mitten'' simplification~\cite{batra2020rearrangement} for grasping that creates a fixed constraint between the hand and the object when they get into contact. We also conduct experiments that remove such simplification for the Fetch robot, in which case the action space includes one additional DoF for grasping. Second, we train agents with a standard imitation learning (IL) algorithm, behavioral cloning~\cite{pomerleau1989alvinn} on \textit{Bimanual Pick and Place}. We collected $30$ demonstrations, more than $6500$ frames in total ($\sim$\SI{215}{\second}). The agent receives ground-truth states of the objects and proprioception information as observation, and outputs the desired linear and angular velocities for both hands. Additional information about the experimental setup can be found in Sec.~\ref{ss:expdetails}.

\textbf{RL experiments:} With the simplified grasping mechanism, the agents trained with SAC for both the bimanual humanoid and the Fetch embodiment achieve $100\%$ success rate for \textit{Grasping Book}, \textit{Soaking Towel}, \textit{Cleaning Stained Shelf}, \textit{Cooking Meat} tasks. For \textit{Slicing Fruit}, the agents achieve only $15\%$ and $0\%$ for bimanual humanoid and Fetch robot respectively, due to the increased accuracy necessary to align the knife blade with the fruit. The agent using bimanual humanoid achieves $0\%$ success in the \textit{Bimanual Pick and Place} task because of the difficulties of controlling and coordinating both hands. Additionally, we evaluate the performance with the Fetch robot in more realistic conditions, without any simplification for grasping, and observe a significant drop in performance, achieving $25\%$ success rate for 2 tasks and $0\%$ for the other 3 (see Fig.~\ref{fig:exp_reward_fetch} in the Appendix). This indicates that successful grasping for diverse objects is a significant challenge in these manipulation tasks. To test generalization, we conducted an ablation study in which we train policies with three different levels of variability in \textit{Soaking Towel} --no variations, different poses, different objects and poses-- and evaluate them on an unseen setup (an unseen object with randomized initial pose).
The policies achieve success rate of $19\%$, $79\%$, and $87\%$, respectively. This study shows that it's essential to train with diverse object models and initial states to obtain robust policies, and the generative system of \ig facilitates it by specifying a few logical states that describe the initial scene.
The attached video shows the policies performing the tasks trained using \ig's new extended physical states and logical predicates that help generating task instances and discriminating their completion.

\textbf{IL experiments in \textit{Bimanual Pick and Place} with VR demonstrations:} 
The trained policy in the full task diverges and fails, even after training with 30 demonstrations. We hypothesize that the agent suffers from covariate shift~\cite{ross2011reduction}, visiting state space not covered by demonstrations, due to the different strategies demonstrated in VR for this long task (300+ steps). We then evaluate if the policy can successfully perform the last part of the task. We take a successful human demonstration and initialize a few seconds before task completion. We then query the IL policy from this point onward to control the agent. The results of this experiment are shown in Fig~\ref{fig:expcurves}. The policy achieves $19\%$ and $46\%$ success rate when starting $\SI{6}{s}$ and $\SI{3}{s}$ away from the goal. This experiment demonstrates the potential of the new VR interface of \ig for generating demonstrations for IL. We believe that our new interface will open new avenues for research in bimanual manipulation, hand-eye coordination, and coordination of robot base and robot arm.

%% file: 5-appendix.tex


\renewcommand{\thesection}{\Alph{section}}
\renewcommand{\thesubsection}{A.\arabic{subsection}}

\setcounter{figure}{0} \renewcommand{\thefigure}{A.\arabic{figure}}

\setcounter{table}{0}
\renewcommand{\thetable}{A.\arabic{table}}

\newpage

\section*{Appendix for \ig: Object-Centric Simulation for Robot Learning of Everyday Household Tasks}
\label{s:appendix}

\subsection{The \ig Virtual Reality Interface}
\label{ss:vrdetails}
In this section, we provide additional information about the implementation of our virtual reality (VR) interface in \ig. 

\textbf{Mapping human motion to the virtual embodiment:} Humans in VR control a bimanual embodiment in \ig composed of a head, two hands and a torso/body. The hardware for VR control is composed of a headset and two hand controllers, possibly an additional tracker for the human torso that maps directly to the VR body. At each step, the pose of the agent's head is directly set to be the new headset's pose as provided by the VR hardware. This step bypasses physics simulation of the head motion to make sure that the motion of the head in VR corresponds exactly and without delays to that in the real world to avoid any discomfort. In contrast, both the hands and the main body move as the result of physical simulation of a body constraint connecting the simulated hands and body to the new poses provided by the VR hardware. In other words, the VR hardware provides at each step new ``desired'' poses for the hands and body that ``pull'' the body parts towards them with forces computed by the simulator. This creates realistic interactions between the hands and body and the objects in the scene, colliding with them and applying forces. The constraint to move the hand has a maximal force of \SI{300}{\newton}, simulating the lifting force of humans. The constraint to move the body is \SI{50}{\newton}. While the desired new poses for the hands always come from the hand-controllers, the desired pose for the body can be provided by a body tracker on the human's torso or be otherwise estimated based on the headset position. This constraint-based system strikes a balance between accurately following the human motion and realistic interactions with the scene in VR.


\textbf{Haptic feedback:} To approximate the real-world experience, it is important to create haptic feedback to human subjects when interacting with the scene. To that end, in our VR interface collisions of the body and the hands trigger haptic vibrations in the controllers. The body will trigger a strong vibration in both controllers facilitating navigation in the scene. The hands also generate low-strength vibration when they are in contact with an object, to notify users that they are in contact with a virtual object. These mechanisms create a multi-modal stream to the humans (vision and haptics) that help them interact more dexterously and realistically.

\textbf{Assistive Grasping:} Creating realistic, robust and dexterous grasping in virtual reality is challenging. Grasping objects in real-world involves generating multiple frictional contact points and surfaces between the hand and the object: simulating physically this complex process in a realistic manner is non-trivial. Additionally, real-world grasping involves rich multimodal signals that include tactile and haptic information, that is not available in common virtual reality interfaces. To compensate for these differences and generate a natural experience in simulation, we implement an assistive grasping mechanism. 

In our VR interface, grasping is performed by pressing the left or right trigger, a one degree of freedom (DoF) actuation. This single DoF is mapped to a closing motion of the avatar's hand where all fingers move synchronously. If the trigger is pressed more than 50\% of its range, we activate the \textit{assistive grasping} (AG) mode. The AG mode facilitates grasping by creating an additional fixed or point-to-point joint between the hand and the movable objects inside the hand. We use three criteria to decide what object inside the hand should be assisted for grasping: First, the object has to be inside the hand. We evaluate this we a ray casting mechanism. Rays are shot between the following two sets of points: (thumb tip, thumb middle, palm middle, palm base) and (4 non-thumb finger tips). These 16 rays return a list of all objects that are in the hand. Second, the object has to be close to the palm. This is defined as the distance between the center of mass of the palm link and the object is the smallest among all objects in hand. And third, the object needs to be in contact with the hand and the hand has to be applying force on it.
We found that, defined with these three criteria, the AG mechanism is realistic as humans can grasp objects reliably with motions that are close to the ones used in real-world for the same objects. 
While a user is grasping an object using AG, collision of the hand with that object is also disabled, to avoid any recurring collisions or simulation instability. 

The AG breaks the connection between the object and the hand if the grasping trigger goes under 50\% pressed, or the center of mass of object being grasped moves further away from the palm than a maximum threshold, $D_\mathit{AGMax}$. We use $D_\mathit{AGMax}=\SI{10}{cm}$. This effectively avoids that humans can use AG to grasp objects or pull from them in an unrealistic manner, e.g., grasping a heavy object such a watermelon with a single hand. Our AG system strikes a good balance between facilitating grasping and interactions of humans in VR and simplifying the manipulation of objects excessively. In this way, humans in VR still need to move and position the hands in realistic ways to grasp and manipulate heavy objects. The AG mode is also critical to grasp small objects with dexterity. In the VR demonstrations created by the human, we can see that the AG mechanism works robustly for all kind of rigid and articulated objects, ranging from fridge handles to knives, from watermelons to strawberries. Please refer to the supplementary video to see examples of AG mechanism. 

\textbf{Hardware compatibility:} We have tested our \ig with the three main commercially available headsets at the time, HTC Vive (Pro Eye), Oculus Rift S and Oculus Quest. For the former, we include functionalities to observe and record the eye gaze tracked by the device. To serve both devices, the iGibson renderer needs to output 1296 x 1440 images at \SI{30}{\hertz}, shown to the human to create a immersive 3D experience. The system is capable of rendering at up to \SI{90}{\hertz}, but we had to reduce the frame rate to accommodate additional expensive per-frame computations, including physics and extended states update steps in our simulator.


\textbf{Logging and replaying demonstrations:} All information of the runs can be logged and replay deterministically (same output for the same actions). The logged information includes kinematics and extended \ig states, and VR agent actions. We believe the logged information acquired with the \ig VR interface will facilitate research: the information can be analyzed to understand human strategies, or used with modern robot learning techniques, e.g. with imitation learning, to train embodied AI solutions.

\subsection{Extended Object States, Logical Predicates and Generative System in \ig}
\label{ss:kinematicsampling}

In this section, we provide additional information on \ig's new extended physical states, logical predicate system that map physical states to logical states, and the generative system that sample valid simulated physical states based on logical states.

\textbf{Extended states associated to object categories:} In \ig, not all extended states need to be maintained for instances of all object categories, e.g., most objects cannot be \texttt{Sliced} and only food objects could be \texttt{Cooked}. We assume that any object instance added to \ig belongs to a category annotated with properties. The properties indicate what extended states should be updated for object instances of that category. The exhaustive list of all possible object category properties is shown in Table~\ref{table:categoryannotation}. 
Some properties will require additional annotation for each object model to estimate logical states.

\textbf{Object model annotations:} Each model needs to be annotated with additional physical and semantic information to simulate correctly interactions and their associated logical states. The exhaustive list of all possible object model properties are included in Table~\ref{table:staticobjectmodelprops}. Some properties directly come from the 3D assets, such as \texttt{Shape} and \texttt{KinematicStructure}. We compute \texttt{Weight} based on query results from Amazon Product API, and compute \texttt{CenterOfMass} and \texttt{MomentOfInertia} accordingly for each link based on the assumption of uniform density. Note that if an object category is annotated with a certain property, e.g., \texttt{stove} is annotated as \texttt{HeatSourceSink}, we need to annotate \texttt{HeatSourceSinkLink}, a virtual (non-colliding) fixed link that heats or cools objects, for all object models of \texttt{stove}. For \texttt{SlicingTool} and \texttt{CleaningTool}, we additionally annotate \texttt{SlicingToolLink} and \texttt{CleaningToolLink}, which are colliding fixed links that can slice objects (the blade of a knife) and remove dirt particles from objects (the bottom of a vacuum), respectively.

\textbf{Updating object state:} During simulation, our simulator maintains and updates not only the kinematic states of the objects, such as \texttt{Pose} and \texttt{InContactObjs}, using our underlying physics engine Bullet~\cite{coumans2016pybullet}, but also the non-kinematic states, such as \texttt{Temperature} and \texttt{WetnessLevel} with custom rules. These update rules are explained in Sec.~\ref{s:extendedstates} and summarized in Table~\ref{table:objectstates}. 

\textbf{Logical predicates as discriminative functions:} As explained in Sec.~\ref{s:ls}, we define a set of discriminative functions that map the extended physical states to logical states that are semantically grounded on natural language, such as \texttt{Cooked} and \texttt{Sliced}. These logical states can used for symbolic planning and checking intermediate success for sub-tasks for reinforcement learning. The details of the discriminative functions of all the logical predicates can be found in Table~\ref{table:supp_obj_states}.

\textbf{Logical predicates as generative functions:} We also define a set of sampling functions that can generate valid physical states that satisfy the given logical states. For example, if the initial conditions of the task require a book placed \texttt{OnTopOf} a table or a shelf being \texttt{Stained}, our system can automatically sample concrete physical states that satisfy the requirements: sampling a random position on the table and place the book there, and sample stain particles on random locations on the shelf (see Fig.~\ref{fig:exptasks}). The details of the generative functions of all the logical predicates can be found in Table~\ref{table:supp_generative_func}. Please also refer to our supplementary video for more details. 

Sampling extended states based on the given logical predicates is relatively simple, e.g., sampling a temperature that corresponds to an object being \texttt{Frozen}. However, sampling object poses to fulfill the given kinematic predicates is more involved as it requires sampling values in the Special Euclidean group SE(3) with additional constraints such as placing objects in stable configurations and not causing penetrations between objects. In the following, we describe our algorithm to sample valid poses based on kinematic logical predicates.

\paragraph{Pose Sampling Algorithm:} Say we are sampling a valid pose for an object $o_1$ to be \texttt{OnTopOf} object $o_2$. First, we query the set of stable orientations allowed for object $o_1$. We assume these orientations are provided per object model, e.g., for a book the orientations to place the book on its cover and last page or upright. Each stable orientation is linked to an axis-aligned bounding box with an associated bounding-box base area. The next step would be to find areas on the surface of object $o_2$ that can hold the bounding-box area and that are flat, unobstructed, and accessible. 
To find these areas of $o_2$ surface we use a ray-casting mechanism conditioned on the specific kinematic logical predicate. For example, for our case of \texttt{OnTopOf} we will generate rays starting immediately above $o_2$ by sampling points from the top face of its axis-aligned bounding box, and marching downwards in the vertical direction.
The points where the rays intersect $o_2$ surface will be used to define planes where we can attempt to sample object $o_1$ if they fulfill some criteria such as providing stable support. We can repeat the procedure for different stable orientations of $o_1$. Other logical predicates use a similar generative procedure but with variations in the ray-tracing step. For example, for \texttt{InsideOf}, we start our rays at different points inside the $o_2$ bounding box rather than above it. In addition, for particle-based states such as \texttt{Dusty} and \texttt{Stained}, we additionally allow casting rays in horizontal directions.

\input{preds_table}

\input{obj_cat_props_table}

\input{object_model_props_table}

\input{object_states_table}

\input{sampling_table}

\FloatBarrier

\subsection{Experimental Setup and Additional Results}
\label{ss:expdetails}
In this section we provide the experimental setup for the reinforcement learning and imitation learning experiments described in Sec.~\ref{s:eval}. 

\paragraph{Reinforcement Learning Experiments with Bimanual Humanoid Robot:} The observation space include $128 \times 128$ RGB-D images from the onboard sensor on the agent's head, and proprioceptive information (hand poses in agent's local frame, and a fraction indicating how much each hand is closed). The action space is 6-dimensional representing the desired linear and angular velocities of the right hand, where the rest of the agent is stationary. For grasping, we adopt the ``sticky mitten'' simplification from other works~\cite{batra2020rearrangement}: we create a fixed constraint between the hand and the object as soon as they get in contact. 

\begin{figure}[t]
\centering
\includegraphics[width=0.32\linewidth]{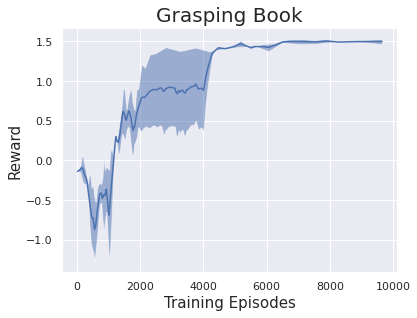}
\includegraphics[width=0.32\linewidth]{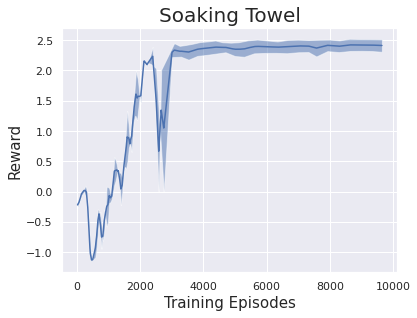}
\includegraphics[width=0.32\linewidth]{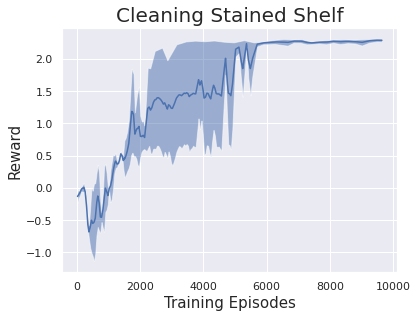}
\\\vspace{0.1cm}
\includegraphics[width=0.32\linewidth]{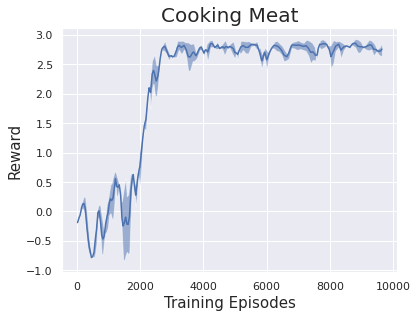}
\includegraphics[width=0.32\linewidth]{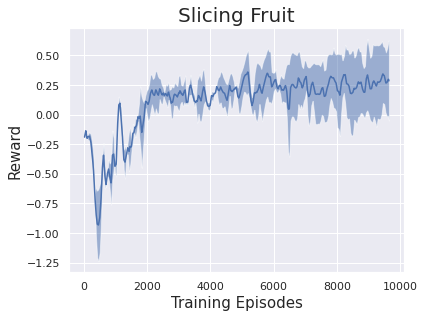}
\includegraphics[width=0.32\linewidth]{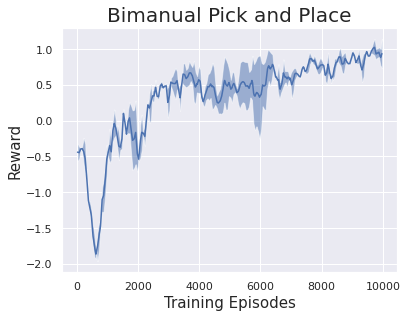}

\caption{\textbf{Reward curves for RL experiments (Bimanual Humanoid)}: We observe steady training progress across all the tasks and the RL agent achieve perfect reward for task 1)-4). For \textit{Slicing Fruit}, the agent achieves only $15\%$ success rate because the task requires a precise alignment between the knife blade and the fruit. For \textit{Bimanual Pick and Place}, the agent fails to succeed; although it receives partial reward for approaching the cauldron, bimanual manipulation of large heavy objects requires careful coordination between hands and remains challenging learn with RL.
}
\label{fig:exp_reward}
\end{figure}

The agent receives a one-time success reward if it satisfies the single predicate (e.g. \texttt{Cooked(meat)}). Additionally, we provide distance-based reward shaping for each experiment to encourage the hand to approach activity-relevant objects, e.g. encourage the hand to approach the meat and the meat to approach to stove. Finally, for the Cleaning Stained Shelf task, we provide partial progress reward for each stain particle that has been cleaned. The episode terminates if the agent achieves success or times out (200 timesteps, or equivalently 20 seconds). We train for 10K episodes, evaluate on the same setups, and report the results. The training reward curves can be found in Fig.~\ref{fig:exp_reward}.

We use Soft Actor-Critic~\cite{haarnoja2018soft} for training. The policy network has two encoders for RGB-D images and proprioceptive information. With RGB-D images as input, we use a 3-layer convolutional neural network to encode the image into a 256 dimensional vector. The proprioceptive information is encoded into a 256 dimensional vector with an MLP. The features are concatenated and pass through additional MLP layers to generate the action.

\begin{figure}[t]
\centering
\includegraphics[width=0.32\linewidth]{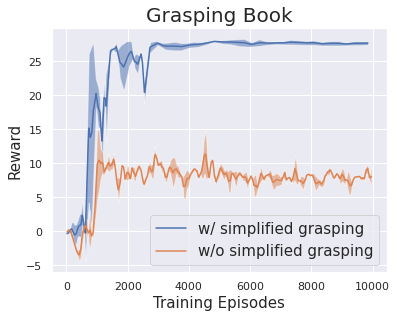}
\includegraphics[width=0.32\linewidth]{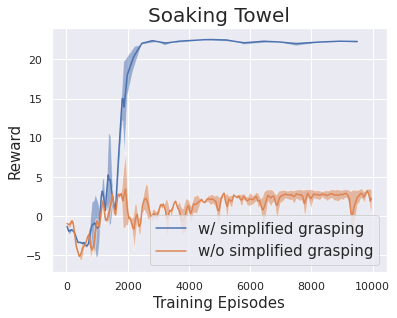}
\includegraphics[width=0.32\linewidth]{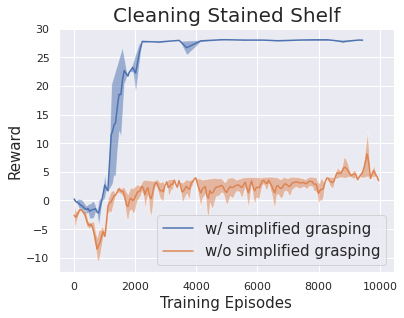}
\\\vspace{0.1cm}
\includegraphics[width=0.32\linewidth]{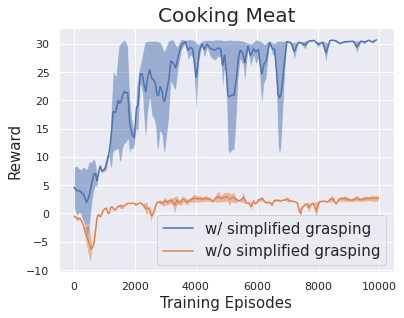}
\includegraphics[width=0.32\linewidth]{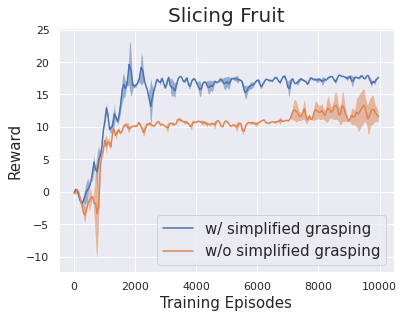}

\caption{\textbf{Reward curves for the five task trained with RL (Fetch)}: When we adopt the same ``sticky mitten'' grasping simplifcation as Bimanual Humanoid for Fetch, the RL agent achieves very similar result (blue) as the one shown in Fig.~\ref{fig:exp_reward}. Once we remove such simplification, the RL agent, however, struggles to solve the tasks, achieving around $25\%$ success rate for \textit{Grasping Book} and \textit{Soaking Towel}, and $0\%$ for the other three. Although the robot learns to approach the task-relevant objects with its end effector, grasping a diverse set of objects (e.g. towel, knife, meat) with a parallel gripper remains a challenging robotics problem.
}
\label{fig:exp_reward_fetch}
\end{figure}

\paragraph{Reinforcement Learning Experiments with Fetch Robot:} We also conducted the same RL experiments with a Fetch robot. The observation space include $128 \times 128$ RGB-D images from the onboard sensor on the agent's head, and proprioceptive information (the end effector pose in agent's local frame, joint configurations, and whether the end effector is currently grasping something). The action space is 6-dimensional representing the desired linear and angular velocities of the end effector, where the rest of the agent is stationary. We experimented with both the ``sticky mitten'' grasping simplification (the same as Bimanual Humanoid) and without such simplification. For the later setup, the Fetch robot has to rely on the friction between the gripper fingers and the objects to grasp them with realistic physics simulation. Its action space also includes one additional DoF for closing the gripper. With this later setup, we hope to minimize the sim2real gap as much as possible. Due to the additional complexity of grasping, we add one more reward shaping terms to encourage the gripper to grasp the task-relevant object and penalize the agent for dropping it. We use different reward scaling. The termination conditions remain the same. We also use the same policy network architecture and training schema as before. The training reward curves can be found in Fig.~\ref{fig:exp_reward_fetch}.

\paragraph{Imitation Learning Experiments with Bimanual Humaniod Robot}:
The observation space includes the ground truth poses of the task-relevant objects, and proprioceptive information, the same as the RL setup. In the \textit{Bimanual Pick and Place} experiment, task relevant objects include the cauldron, the table and the agent itself. The agent can control both of its hands with the desired linear and angular velocities, which result in 12 degree of freedom. The hand closing action is not learned, but replayed from the human demonstrations. 

We collected $6500+$ state-action pairs from $30$ human demos and used behavior cloning to predict the action based on the state. The policy network has two MLP encoders for proprioceptive information and ground truth object poses. The features are concatenated and pass through additional MLP layers to generate the action. 

\begin{figure}[!t]
    \centering
    \includegraphics[width=0.23\linewidth]{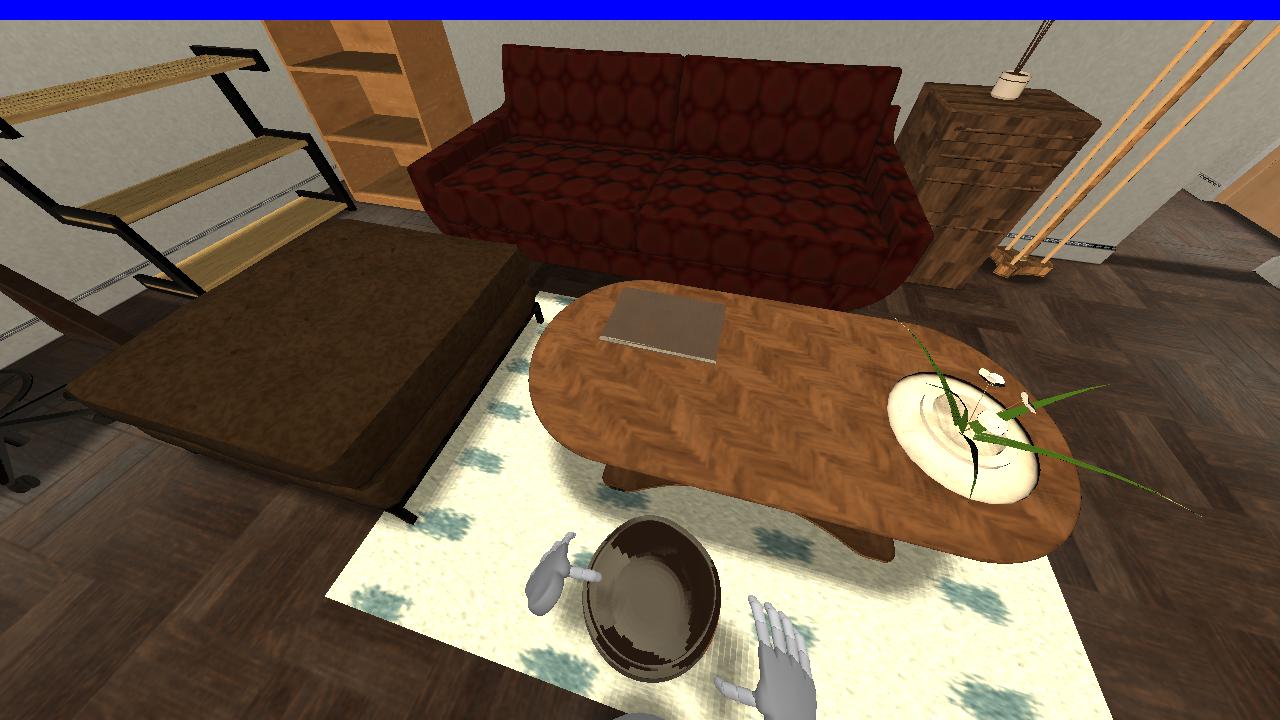}
    \hfill
    \includegraphics[width=0.23\linewidth]{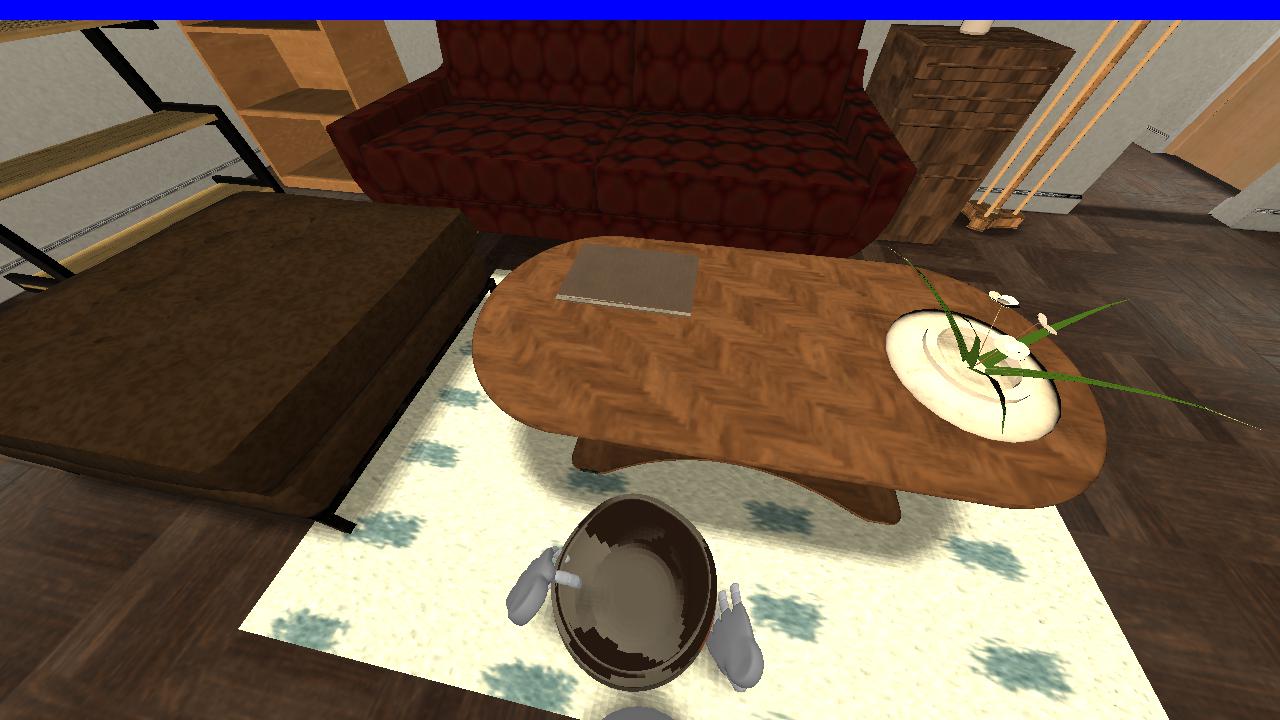}
    \hfill
    \includegraphics[width=0.23\linewidth]{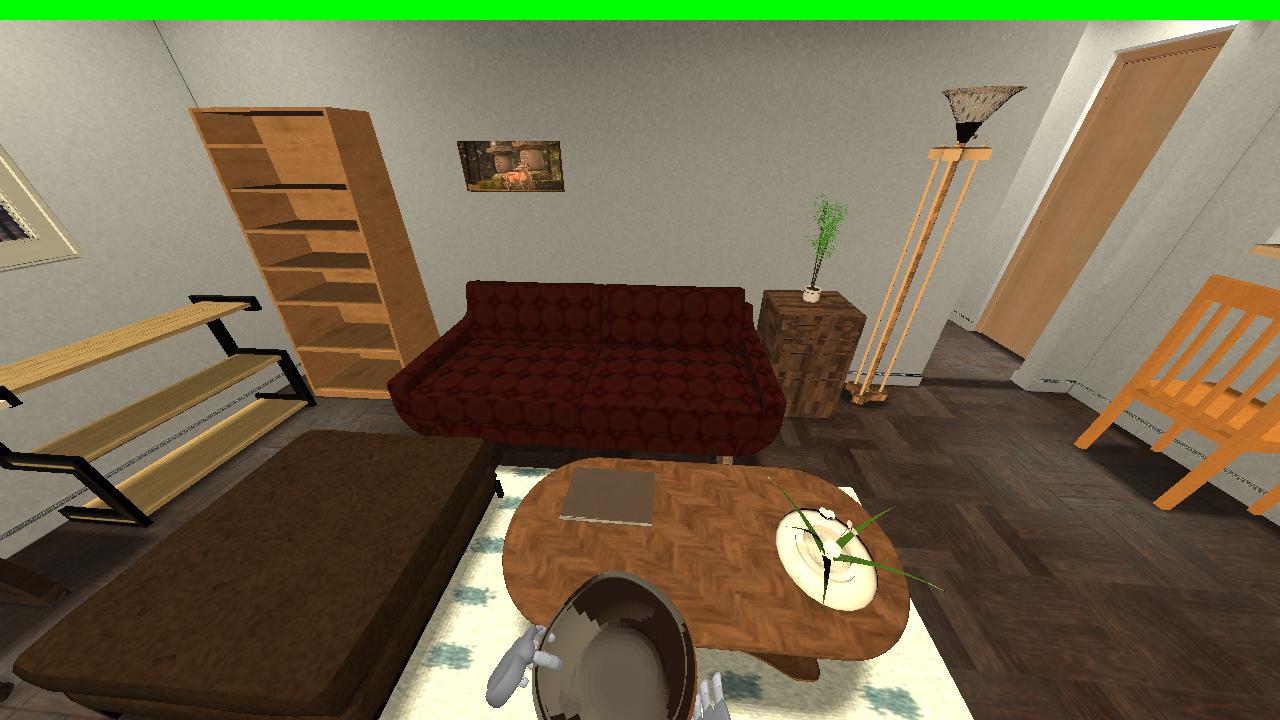}
    \hfill
    \includegraphics[width=0.23\linewidth]{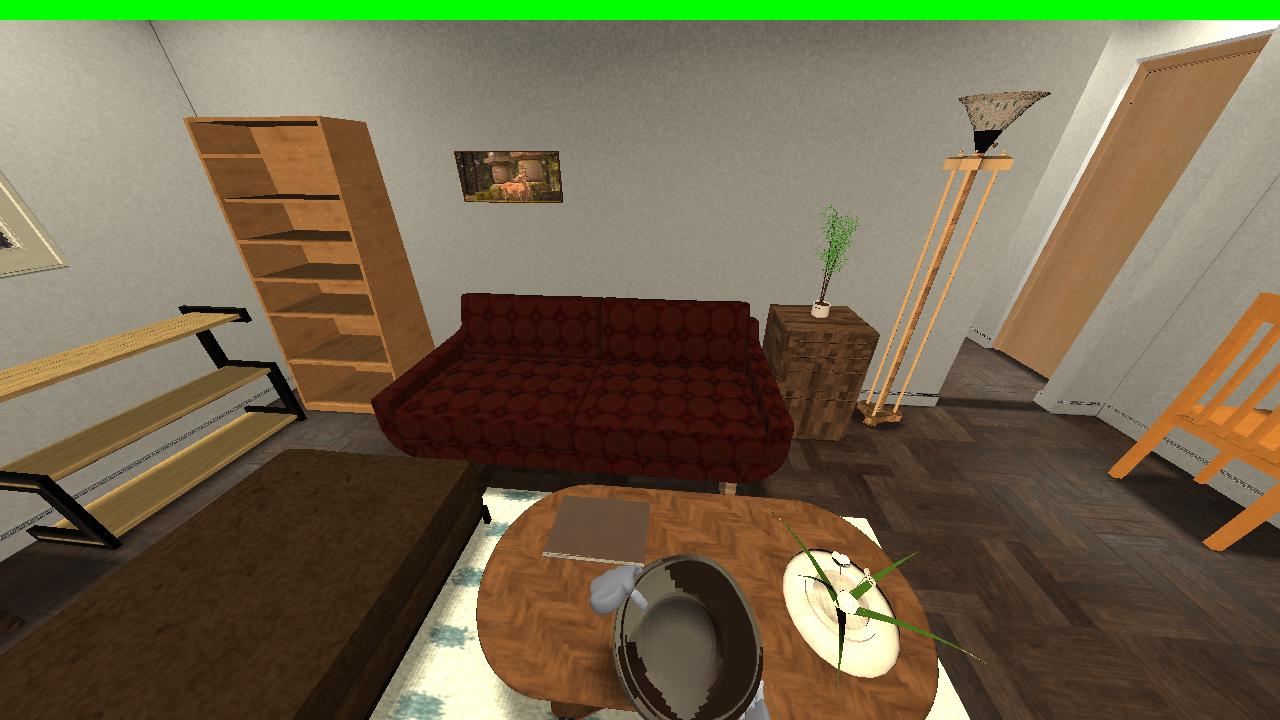}
    \caption{\textbf{A sequence of IL execution of bimnual grasping}: the blue bar indicates that we are replaying the demonstration, and the green bar is indicating the policy is taking over. The sequence shows that we are able to complete the task of placing a heavy object on table by starting at $T-2\text{s}$. }
    \label{fig:il_sequence}
\end{figure}

The network is trained until validation loss plateaus, and evaluated on a test set of demonstrations. As discussed in the main paper, the entire task is long horizon (>300 steps) and the policy diverges due to covariate shift~\cite{ross2011reduction}. We then evaluate if the policy can successfully perform the task if we initialize the simulator a few seconds before task completion of a successful human demo. The success rate with respect to different policy starting time can be found in Fig.~\ref{fig:expcurves}. We show a successful sequence after rewinding 2 seconds in Fig.~\ref{fig:il_sequence}.

\subsection{Performance Benchmark of \ig}

To evaluate whether \ig can be used in computationally expensive embodied AI research, we benchmarked the performance (simulation time) and compared with the previous version. The benchmark setup is the same as in \citet{shen2020igibson}, which considered an ``idle'' setup, in which we place a robot (a TurtleBot model) in the scene and run the physics simulation and extended physical state simulation loop. The benchmark runs on 15 scenes, and statistics are collected. The agent applies zero actions and stays still. We use action time step of $t_a = \frac{1}{30}\text{s}$ and physics time step of $t_s = \frac{1}{120}\text{s}$ to be consistent with \citet{shen2020igibson}. Both settings are benchmarked on a computer with Intel 5930k CPU and Nvidia GTX 1080 Ti GPU, in a single process setting, rendering $512\times512$ RGB-D images.

The simulator speed is shown in Table~\ref{tbl:sim_speed}. Although we added many extended physical states, we still achieved a $25\%$ increase in average performance compared with iGibson 1.0~\cite{shen2020igibson}. In \ig, the main source of speed up with respect to the previous version of iGibson is obtained from better usage of the object sleeping mechanism, and lazy update of object poses in the renderer. This allows us to simulate much larger scenes with many more objects with extended physical states tracked, and as a result, more diverse everyday household activities.

\begin{table}[t]
  \begin{center}
  \centering
  \begin{tabular}{lcccc} 
  \toprule
  Simulator &  Mean & Max & Min \\
  \midrule
    iGibson 1.0 & 100 & 150 & 68 \\
  \midrule
 iGibson 2.0 & 125 & 217 & 73 \\
   \bottomrule
  \end{tabular}
  \end{center}
  \vspace{3pt}
    \caption{Simulation Speed Comparison for iGibson 1.0 and iGibson 2.0. The unit is steps per second; each step simulates $\frac{1}{30}\text{s}$. \label{tbl:sim_speed}}
\end{table}

\subsection{Feature Comparisons of Simulators}
In Table~\ref{t:simenv}, we provide a detailed comparison across multiple simulation environments. The table is adapted from Table I of ~\cite{shen2020igibson}. We include more recent simulation environments as columns and more feature comparisons as rows.
\input{comparison_table}

\subsection{Limitations and Future Work}
Although \ig has made several significant contributions towards simulating complex, everyday household tasks for robot learning, it is not without limitation. First of all, \ig doesn't support soft bodies / flexible material in a scalable way at the moment, due to the limitation of our underlying physics engine. This prevents us from simulating tasks like folding laundry and making bed in large, interactive scenes. Also, \ig doesn't support accurate human behavior modeling (other than goal-oriented navigation), and thus prevent us from simulating tasks that are inherently rich in human-robot interaction (e.g. elderly care). With the recent advancement of physics engines, and human behavior modeling and motion synthesis, we plan to overcome these limitations in the future. In addition, we also plan to support a more diverse set of extended object states (e.g. \texttt{Filled}, \texttt{Hung}, \texttt{Assembled}, etc) as well as bi-directional transitions for some of our existing states (e.g. \texttt{Soaked} and \texttt{Stained}/\texttt{Dusty}), which can unlock even more household tasks. Finally, we plan to transfer mobile manipulation policies trained in \ig to the real world.

%% file: preds_table.tex
\begin{table}[!htbp]
\begin{center}
{\scriptsize%
\renewcommand{\arraystretch}{1.2}
\begin{tabular}{ | m{2.5cm} | m{10.5cm} | } 
\hline
\textbf{Predicate} & \textbf{Description} \\ 
\hline
\texttt{InsideOf($o_1$,$o_2$)} &  Object $o_1$ is inside of object $o_2$ if we can find two orthogonal axes crossing at $o_1$ center of mass that intersect $o_2$ collision mesh in both directions.\\ 
\hline
\texttt{OnTopOf($o_1$,$o_2$)} &  Object $o_1$ is on top of object $o_2$ if $o_2 \in \texttt{InSameNegativeVerticalAxisObjs}(o_1) \land o_2 \not\in \texttt{InSamePositiveVerticalAxisObjs}(o_1) \land \texttt{InContactWith}(o_1, o_2)$, where $\texttt{InSamePositive/NegativeVerticalAxisObjs}(o_1)$ is the list of objects in the same positive/negative vertical axis as $o_1$ and $\texttt{InContactWith}(o_1, o_2)$ is whether the two objects are in physical contact.\\ 
\hline
\texttt{NextTo($o_1$,$o_2$)} &  Object $o_1$ is next to object $o_2$ if $o_2 \in \texttt{InSameHorizontalPlaneObjs}(o_1) \land l2(o_1, o_2) < t_\mathit{NextTo}$, where $\texttt{InSameHorizontalPlaneObjs}(o_1)$ is a list of objects in the same horizontal plane as $o_1$, $l2$ is the L2 distance between the closest points of the two objects, and $t_\mathit{NextTo}$ is a distance threshold that is proportional to the average size of the two objects. \\
\hline
\texttt{InContactWith($o_1$,$o_2$)} &  Object $o_1$ is in contact with $o_2$ if their surfaces are in contact in at least one point, i.e., $o_2 \in \texttt{InContactObjs}(o_1)$. \\ 
\hline
\texttt{Under($o_1$,$o_2$)} &  Object $o_1$ is under object $o_2$ if $o_2 \in \texttt{InSamePositiveVerticalAxisObjs}(o_1)$ $\land o_2 \not\in \texttt{InSameNegativeVerticalAxisObjs}(o_1)$.\\
\hline
\texttt{OnFloor($o_1$,$o_2$)} &  Object $o_1$ is on the room floor $o_2$ if $\texttt{InContactWith}(o_1, o_2)$ and $o_2$ is of \texttt{Room} type. \\ 
\hline
\texttt{Open(o)} &  Any joints (internal articulated degrees of freedom) of object $o$ are open. Only joints that are relevant to consider an object \texttt{Open} are used in the predicate computation, e.g. the door of a microwave but not the buttons and controls. To select the relevant joints, object models of categories that can be \texttt{Open} undergo an additional annotation that produces a \texttt{RelevantJoints} list. A joint is considered open if its joint state $q$ is 5\% over the lower limit, i.e. $q > 0.05(q_\mathit{UpperLimit} - q_\mathit{LowerLimit}) + q_\mathit{LowerLimit}$. \\
\hline
\texttt{Cooked(o)} &  The temperature of object $o$ was over the cooked threshold, $T_\mathit{cooked}$, and under the burnt threshold, $T_\mathit{burnt}$, at least once in the history of the simulation episode, i.e., $T_\mathit{cooked} \leq T_o^\mathit{max} < T_\mathit{burnt}$. We annotate the cooked temperature $T_\mathit{cooked}$ for each object category that can be \texttt{Cooked}.\\
\hline
\texttt{Burnt(o)} &  The temperature of object $o$ was over the burnt threshold, $T_\mathit{burnt}$, at least once in the history of the simulation episode, i.e., $T_o^\mathit{max} \geq T_\mathit{burnt}$. We annotate the burnt temperature $T_\mathit{burnt}$ for each object category that can be \texttt{Burnt}.\\
\hline
\texttt{Frozen(o)} &  The  temperature of object $o$ is under the freezing threshold, $T_\mathit{frozen}$, i.e., $T_o \leq T_\mathit{frozen}$. We assume as default freezing temperature $T_\mathit{frozen}=0^\circ C$, a value that can be adapted per object category and model.\\
\hline
\texttt{Soaked(o)} &  The  wetness level $w$ of the object $o$ is over a threshold, $w_\mathit{soaked}$, i.e., $w \geq w_\mathit{soaked}$. The default value for the threshold is $w_\mathit{soaked}=1$, (the object is soaked if it absorbs one or more droplets), a value that can be adapted per object category and model.\\
\hline
\texttt{Dusty(o)} &  The  dustiness level $d$ of the object $o$ is over a threshold, $d_\mathit{dusty}$, i.e., $d > d_\mathit{dusty}$. The default value for the threshold is $d_\mathit{dusty}=0.5$, (half of the dust particles have been cleaned), a value that can be adapted per object category and model.\\
\hline
\texttt{Stained(o)} &   The  stain level $s$ of the object $o$ is over a threshold, $s_\mathit{stained}$, i.e., $s > s_\mathit{stained}$. The default value for the threshold is $s_\mathit{stained}=0.5$, (half of the stain particles have been cleaned), a value that can be adapted per object category and model.\\
\hline
\texttt{ToggledOn(o)} & Object $o$ is toggled on or off. It is a direct query of the \ig objects' extended state $\mathit{TS}$, the toggled state.\\ 
\hline
\texttt{Sliced(o)} & Object $o$ is sliced or not. It is a direct access of the \ig objects' extended state $\mathit{SS}$, the sliced state.\\
\hline
\texttt{InFoVOfAgent(o)} &  Object $o$ is in the field of view of the agent, i.e., at least one pixel of the image acquired by the agent's onboard sensors corresponds to the surface of $o$.\\
\hline
\texttt{InHandOfAgent(o)} &  Object $o$ is grasped by the agent's hands (i.e. assistive grasping is activated on that object).\\
\hline
\texttt{InReachOfAgent(o)} &  Object $o$ is within $d_\mathit{reach}=2$ meters away from the agent.\\
\hline
\texttt{InSameRoomAsAgent(o)} &  Object $o$ is located in the same room as the agent.\\
\hline
\end{tabular}
}
\vspace{3pt}
\caption{\textbf{Logical Predicates:} Description of the discriminative functions}
\label{table:supp_obj_states}
\end{center}
\end{table}

%% file: obj_cat_props_table.tex
\begin{table}[!htbp]
\begin{center}
{%
\scriptsize
\renewcommand{\arraystretch}{1.2}
\begin{tabular}{ | m{3.5cm} | m{3.5cm} | } 
\hline
\textbf{Property of an object category} & \textbf{Required extended object states}\\ 
\hline
Can be \texttt{Cooked} & \texttt{MaxTemperature}, \texttt{Temperature}\\ 
\hline
Can be \texttt{Burnt} & \texttt{MaxTemperature}, \texttt{Temperature}\\ 
\hline
Can be \texttt{Frozen} & \texttt{Temperature}\\ 
\hline
Can be \texttt{Soaked} & \texttt{WetnessLevel}\\ 
\hline
Can be \texttt{Dusty} & \texttt{DustinessLevel}\\ 
\hline
Can be \texttt{Stained} & \texttt{StainLevel}\\ 
\hline
Can be \texttt{ToggledOn} & \texttt{ToggledState}\\ 
\hline
Can be \texttt{Sliced} & \texttt{SlicedState}\\ 
\hline
Is a \texttt{HeatSourceSink} & \texttt{ToggledState}\\ 
\hline
Is a \texttt{DropletSource} & \texttt{ToggledState}\\ 
\hline
\end{tabular}
}
\vspace{3pt}
\caption{Extended states associated to properties of object categories}
\label{table:categoryannotation}
\end{center}
\end{table}

%% file: object_model_props_table.tex
\begin{table}[!htbp]
\begin{center}
{%
\scriptsize
\renewcommand{\arraystretch}{1.2}
\begin{tabular}{ | m{2.5cm} | m{2.5cm} | m{7.5cm} | } 
\hline
\textbf{Object Model Property} & \textbf{Must be defined if the object {\mbox category\ldots}} & \textbf{Description} \\ 
\hline
\texttt{Shape} &  & Model of the 3D shape of each link of the object\\ 
\hline
\texttt{Weight} &  & Weight of the object \\ 
\hline
\texttt{CenterOfMass} &  & Mean position of the matter in the object\\ 
\hline
\texttt{MomentOfInertia} &  & Resistance of the object to change its angular velocity\\ 
\hline
\texttt{KinematicStructure} &  & Structure of links and joints connecting them in the form of URDF (non-articulated objects are composed of one link)\\ 
\hline
\texttt{StableOrientations} &  & A list of stable orientations assuming the object is placed on a flat surface, computed using a 3D geometry library\\ 
\hline
\texttt{HeatSourceSinkLink} & Is a \texttt{HeatSourceSink} & Virtual (non-colliding) fixed link that generates/absorbs heat\\ 
\hline
\texttt{CleaningToolLink} & Is a \texttt{CleaningTool} & Fixed link that needs to contact dirt particles for the tool to clean them\\ 
\hline
\texttt{DropletSourceLink} & Is a \texttt{DropletSource} & Virtual (non-colliding) fixed link that generates droplets\\
\hline
\texttt{DropletSinkLink} & Is a \texttt{DropletSink} & Virtual (non-colliding) fixed link that absorbs droplets\\ 
\hline
\texttt{TogglingLink} & Can be \texttt{ToggledOn} & Virtual (non-colliding) fixed link that changes the toggled state of the object when contacted\\ 
\hline
\texttt{SlicingLink} & Is a \texttt{SlicingTool} & Fixed link that changes the sliced state of another object if it contacts it with enough force\\ 
\hline
\texttt{RelevantJoints} & Can be \texttt{Open} & List of joints that are relevant to indicate whether an object is open\\ 
\hline
\end{tabular}
}
\vspace{3pt}
\caption{Non-updatable object model properties (annotated)}
\label{table:staticobjectmodelprops}
\end{center}
\end{table}

%% file: object_states_table.tex
\begin{table}[!htbp]
\begin{center}
{\scriptsize%
\renewcommand{\arraystretch}{1.2}
\begin{tabular}{ | m{4cm} | m{9cm} | } 
\hline
\textbf{Object State} & \textbf{Description and Update Rules} \\ 
\hline
\texttt{Pose} & 6 DoF pose (position and orientation) of the object in world reference frame, updated by the underlying physics engine.\\ 
\hline
\texttt{AABB} & Axis-aligned bounding box (coordinates of two opposite corners) of the object in the world reference frame, updated by the underlying physics engine.\\ 
\hline
\texttt{JointStates} & State of all internal DoFs of the (articulated) object for the structure defined by \texttt{KinematicStructure}, updated by the underlying physics engine.\\ 
\hline
\texttt{InContactObjs} & List of all objects in physical contact with the object, updated by the underlying physics engine.\\ 
\hline
\texttt{InSamePositiveVerticalAxisObjs} & List of all objects in the positive vertical axis drawn from the object's center of mass, updated by shooting a ray upwards in the positive z-axis and gather the objects hit by the ray.\\ 
\hline
\texttt{InSameNegativeVerticalAxisObjs} & List of all objects in the negative vertical axis drawn from the object's center of mass, updated by shooting a ray downwards in the negative z-axis and gather the objects hit by the ray.\\ 
\hline
\texttt{InSameHorizontalPlaneObjs} & List of all objects in the horizontal plane drawn from the object's center of mass, updated by shooting a number of ray in the x-y plane and gather the objects hit by the rays.\\ 
\hline
\texttt{Temperature}, $T$ & Object's current temperature in $^\circ$C, updated by detecting if the object is affected by any heat source or heat sink.\\ 
\hline
\texttt{MaxTemperature}, $T_\textit{max}$ & Maximum temperature of the object reached historically during this simulation run, updated by keeping track of all the \texttt{Temperature} in the history.\\ 
\hline
\texttt{WetnessLevel}, $w$ & Amount of liquid absorbed by the object corresponding to the number of liquid droplets contacted, updated by detecting if the object is in contact with any liquid droplets.\\ 
\hline
\texttt{DustinessLevel}, $d$ & Fraction of the initial amount of dust particles that remain on the object's surface, updated by detecting if the particles are in contact with any \texttt{CleaningTool}.\\ 
\hline
\texttt{StainLevel}, $s$ & Fraction of the initial amount of stain particles that remain on the object's surface, updated by detecting if the particles are in contact with any \texttt{Soaked} \texttt{CleaningTool}.\\ 
\hline
\texttt{ToggledState}, $\textit{TS}$ & Binary state indicating if the object is currently on or off, updated by detecting if the agent is in contact with the \texttt{TogglingLink}.\\ 
\hline
\texttt{SlicedState}, $\textit{SS}$ & Binary state indicating whether the object has been sliced (irreversible), updated by detecting if the object is in contact with any \texttt{CleaningTool} that exerts a force above a certain threshold $F_\mathit{sliced}$. We assume as default force threshold of $F_\mathit{sliced} = \SI{10}{\newton}$, a value that can be configured per object category and model.\\ 
\hline
\end{tabular}
}
\vspace{3pt}
\caption{Object states maintained by \ig}
\label{table:objectstates}
\end{center}
\end{table}

%% file: sampling_table.tex
\begin{table}[!t]
\begin{center}
{\scriptsize%
\renewcommand{\arraystretch}{1.2}
\begin{tabular}{ | m{0.20\textwidth} | m{0.7\textwidth} | } 
\hline
\textbf{Predicate} & \textbf{Sampling Mechanism} \\ 
\hline
\texttt{InsideOf($o_1$,$o_2$)} & Only \texttt{InsideOf($o_1$,$o_2$) = True} can be sampled. $o_1$ is randomly sampled within $o_2$ using the mechanism described in Pose Sampling Algorithm in Sec.~\ref{ss:kinematicsampling}. $o_1$ is guaranteed to be supported fully by a surface and free of collisions with any other object except $o_2$.\\ 
\hline
\texttt{OnTopOf($o_1$,$o_2$)} & Only \texttt{OnTopOf($o_1$,$o_2$) = True} can be sampled. $o_1$ is randomly sampled on top of $o_2$ using the mechanism described in Pose Sampling Algorithm in Sec.~\ref{ss:kinematicsampling}. $o_1$ is guaranteed to be supported fully by a surface and free of collisions with any other object except $o_2$.\\ 
\hline
\texttt{NextTo($o_1$,$o_2$)} & Not supported at the moment. \\
\hline
\texttt{InContactWith($o_1$,$o_2$)} & Not supported at the moment. \\ 
\hline
\texttt{Under($o_1$,$o_2$)} & Only \texttt{Under($o_1$,$o_2$) = True} can be sampled. $o_1$ is randomly sampled on top of the floor region beneath $o_2$ using the mechanism described in Pose Sampling Algorithm in Sec.~\ref{ss:kinematicsampling}. $o_1$ is guaranteed to be supported fully by a surface and free of collisions with any other object except the floor.\\
\hline
\texttt{OnFloor($o_1$,$o_2$)} & Only \texttt{OnFloor($o_1$,$o_2$) = True} can be sampled. $o_1$ is randomly sampled on top of $o_2$, which is the floor of a certain room, using the scene's room segmentation mask. $o_1$ is guaranteed to be supported fully by a surface and free of collisions with any other object except $o_2$.\\ 
\hline
\texttt{Open(o)} & To sample an object $o$ with the predicate \texttt{Open(o) = True}, a subset of the object's relevant joints (using the \texttt{RelevantJoints} model property) are selected, and each selected joint is moved to a uniformly random position between the openness threshold and the joint's upper limit. To sample an object $o$ with the predicate \texttt{Open(o) = False}, all of the object's relevant joints (using the \texttt{RelevantJoints} model property) are moved to a uniformly random position between the joint's lower limit and the openness threshold.\\
\hline
\texttt{Cooked(o)} & To sample an object $o$ with the predicate \texttt{Cooked(o) = True}, the object's \texttt{MaxTemperature} is updated to $\max(T_o^\mathit{max}, T_\mathit{cooked})$. Similarly, to sample an object $o$ with the predicate \texttt{Cooked(o) = False}, the object's \texttt{MaxTemperature} is updated to $\min(T_o^\mathit{max}, T_\mathit{cooked} - 1)$. \\
\hline
\texttt{Burnt(o)} & To sample an object $o$ with the predicate \texttt{Burnt(o) = True}, the object's \texttt{MaxTemperature} is updated to $\max(T_o^\mathit{max}, T_\mathit{burnt})$. Similarly, to sample an object $o$ with the predicate \texttt{Cooked(o) = False}, the object's \texttt{MaxTemperature} is updated to $\min(T_o^\mathit{max}, T_\mathit{burnt} - 1)$.\\
\hline
\texttt{Frozen(o)} & To sample an object $o$ with the predicate \texttt{Frozen(o) = True}, the object's \texttt{Temperature} is updated to a uniformly random temperature between $T_\mathit{frozen} - 10$ and $T_\mathit{frozen} - 50$. To sample an object $o$ with the predicate \texttt{Frozen(o) = False}, the object's \texttt{Temperature} is updated to $T_\mathit{frozen} + 1$. \\
\hline
\texttt{Soaked(o)} & To sample an object $o$ with the predicate \texttt{Soaked(o) = True}, the object's \texttt{WetnessLevel} $w$ is updated to match the \texttt{Soaked} threshold of $w_\mathit{soaked}$. To sample an object $o$ with the predicate \texttt{Soaked(o) = False}, the object's \texttt{WetnessLevel} $w$ is updated to 0.\\
\hline
\texttt{Dusty(o)} & To sample an object with \texttt{Dusty(o) = True}, a fixed number (currently 20) of dust particles are randomly placed on the surface of $o$ using the mechanism described in Pose Sampling Algorithm in Sec.~\ref{ss:kinematicsampling}. To sample an object with \texttt{Dusty(o) = False}, all dust particles on the object are removed.\\
\hline
\texttt{Stained(o)} & To sample an object with \texttt{Stained(o) = True}, a fixed number (currently 20) of stain particles are randomly placed on the surface of $o$ using the mechanism described in Pose Sampling ALgorithm in Sec.~\ref{ss:kinematicsampling}. To sample an object with \texttt{Stained(o) = False}, all stain particles on the object are removed.\\
\hline
\texttt{ToggledOn(o)} & The \texttt{ToggledState} of the object is updated to match the required predicate value. \\ 
\hline
\texttt{Sliced(o)} & The \texttt{SlicedState} of the object is updated to match the required predicate value. Also, the whole object are replaced with the two halves, that will be placed at the same location and inherit the extended states from the whole object (e.g. \texttt{Temperature}).\\
\hline
\end{tabular}
}
\vspace{3pt}
\caption{\textbf{Logical Predicates:} Description of the generative functions}
\label{table:supp_generative_func}
\end{center}
\end{table}

%% file: comparison_table.tex
\renewcommand{\arraystretch}{1.5}
\begin{table*}[th!]
\centering
\caption{Comparison of Simulation Environments}
\label{t:simenv}
\rowcolors{2}{gray!25}{white}
    \resizebox{\textwidth}{!}{

  \begin{tabular}{l|c|cccccccc} 
  \toprule
     &  iGibson 2.0 (ours) & iGibson 1.0~\cite{shen2020igibson} & Gibson~\cite{xia2018gibson}& Habitat~\cite{habitat19arxiv} & Sapien~\cite{xiang2020sapien} & AI2Thor~\cite{kolve2017ai2} & ManipulaTHOR~\cite{ehsani2021manipulathor} & VirtualH~\cite{puig2018virtualhome} & TDW~\cite{gan2020threedworld, gan2021threedworld}  \\ 
  \midrule
  \begin{tabular}{@{}c@{}}\textbf{Provided Large Scenes} \\
  Real-World / Designed\end{tabular}
 & \begin{tabular}{@{}c@{}} 15 homes \\(108 rooms)   /  -- \end{tabular} 
 & \begin{tabular}{@{}c@{}} 15 homes \\(108 rooms)   /  -- \end{tabular} 
 & 1400 / -- & -- & -- & -- / 120 rooms & -- / 30 rooms & -- / 7 & -- / 25  \\

 \textbf{Provided Objects} & 1217 (*) & 570 & -- & -- & 2346 & 609 & 150 & 308 & 200 \\
 
  \textbf{Continuous Extended States} & \textcolor{indiagreen}\faCheck & \textcolor{red}\faTimes & \textcolor{red}\faTimes & \textcolor{red}\faTimes & \textcolor{red}\faTimes & \textcolor{red}\faTimes & \textcolor{red}\faTimes & \textcolor{red}\faTimes & \textcolor{red}\faTimes \\
  
 \textbf{Non-kinematic States} & \textcolor{indiagreen}\faCheck & \textcolor{red}\faTimes & \textcolor{red}\faTimes & \textcolor{red}\faTimes & \textcolor{red}\faTimes & \textcolor{indiagreen}\faCheck & \textcolor{red}\faTimes & \textcolor{indiagreen}\faCheck & \textcolor{red}\faTimes \\

   \begin{tabular}{@{}c@{}}\textbf{Geometric Sampling} \\
  Based on Logical States\end{tabular} & \textcolor{indiagreen}\faCheck & \textcolor{red}\faTimes & \textcolor{red}\faTimes & \textcolor{red}\faTimes & \textcolor{red}\faTimes & \textcolor{indiagreen}\faCheck & \textcolor{red}\faTimes & \textcolor{red}\faTimes & \textcolor{red}\faTimes \\
 
 \textbf{Human Interface} 
&\begin{tabular}{@{}c@{}}Mouse, VR\end{tabular} 
&\begin{tabular}{@{}c@{}}Mouse\end{tabular} 
& -
&\begin{tabular}{@{}c@{}}Mouse\end{tabular} 
& - 
&\begin{tabular}{@{}c@{}}Mouse\end{tabular} 
&\begin{tabular}{@{}c@{}}Mouse\end{tabular} 
&\begin{tabular}{@{}c@{}}Natural\\Language  \end{tabular} 
&\begin{tabular}{@{}c@{}}VR\end{tabular}   \\
 
\begin{tabular}{@{}c@{}}\textbf{Agent/World Interaction} \\ \textbf{F}orces, \textbf{P}redefined \textbf{A}ctions \end{tabular}
 & F
 & F & -- & -- & F & F \& PA  & F \& PA & F \& PA & F \\
\begin{tabular}{@{}c@{}}\textbf{Physics Engine}\end{tabular}
& Pybullet
& Pybullet & Pybullet & Bullet & PhysX & Unity & Unity & Unity & Unity \& Flex  \\
%
%
\begin{tabular}{@{}c@{}}\textbf{Supported Task}
\end{tabular}
& \begin{tabular}{@{}c@{}}Nav.\&Manip.\end{tabular}
& \begin{tabular}{@{}c@{}}Nav.\&Manip.\end{tabular} & \begin{tabular}{@{}c@{}}Nav.\end{tabular} 
& \begin{tabular}{@{}c@{}}Nav.\end{tabular}
& \begin{tabular}{@{}c@{}}Nav.\&Manip.\end{tabular}
&
\begin{tabular}{@{}c@{}}
\begin{tabular}{@{}c@{}}Nav.\&Manip. \end{tabular}
 \end{tabular}  
&
\begin{tabular}{@{}c@{}}
\begin{tabular}{@{}c@{}}Nav.\&Manip. \end{tabular}
 \end{tabular}  
& \begin{tabular}{@{}c@{}} Nav.\&Manip. \end{tabular} 
& \begin{tabular}{@{}c@{}} Nav.\&Manip.\end{tabular}  \\
%
\textbf{Speed} & ++
& ++& +& +++ & ++(PBR)/-(RTX) & + & + &  + & +  \\
\begin{tabular}{@{}c@{}}\textbf{Integrated} \textbf{Motion Planner}\end{tabular}
 & \textcolor{indiagreen}\faCheck
 & \textcolor{indiagreen}\faCheck & \textcolor{red}\faTimes & \textcolor{red}\faTimes & \textcolor{red}\faTimes & \textcolor{red}\faTimes & \textcolor{red}\faTimes & \textcolor{red}\faTimes & \textcolor{indiagreen}\faCheck  \\

  \midrule

  \rowcolor{gray!0}
\textbf{Specialty} & 
\begin{tabular}{@{}c@{}}Continuous Extended\\States, VR\end{tabular} &
\begin{tabular}{@{}c@{}}Phys. Int. in\\Large Scenes\end{tabular} &
\begin{tabular}{@{}c@{}}Nav.\end{tabular} & 
\begin{tabular}{@{}c@{}}Fast, \\ Nav. \end{tabular}& 
\begin{tabular}{@{}c@{}}Articulation,\\ Ray Tracing\end{tabular} & 
\begin{tabular}{@{}c@{}}Object States,\\ Task Planning \end{tabular} & 
\begin{tabular}{@{}c@{}}Mobile\\Manipulation \end{tabular} & 
\begin{tabular}{@{}c@{}}Object States,\\Task Planning \end{tabular} & 
\begin{tabular}{@{}c@{}}Audio,\\Fluids \end{tabular} \\


\bottomrule
%
%
%
%
\rowcolor{gray!0}
\multicolumn{8}{l}{\small \textbf{Type of rendering:} \textbf{PBR}:Physics-Based Rendering, \textbf{IBR}:Image-Based Rendering, \textbf{RTX}:Ray Tracing }\\
  \rowcolor{gray!0}
\multicolumn{8}{l}{\small \textbf{Virtual sensor signals:} \textbf{RGB}: Color Images, \textbf{D}:Depth, \textbf{N}:Normals, \textbf{SS}:Semantic Segmentation, \textbf{LiDAR}:Lidar, \textbf{FL}:Flow (optical and/or scene), \textbf{S}: Sounds}\\

\rowcolor{gray!0}
\multicolumn{8}{l}{\small * included in a parallel submission BEHAVIOR and fully compatible with \ig}\\

%
%
\end{tabular}
}
\end{table*}

    
